\documentclass[onefignum,onetabnum]{siamonline190516}
\usepackage{svg}
\usepackage[utf8]{inputenc} 
\usepackage[T1]{fontenc}    
\usepackage{hyperref}       
\usepackage{url}            
\usepackage{booktabs}       
\usepackage{amsfonts}       
\usepackage{nicefrac}       
\usepackage{microtype}      
\usepackage{bm} 
\usepackage{amsmath,amssymb}
\usepackage{wrapfig}
\usepackage{graphicx}
\graphicspath{{./figures/}}
\usepackage{subcaption}
\usepackage{enumitem}
\usepackage{tikz}  
\usepackage{soul} 
\usepackage{multicol}
\usepackage{multirow}
\usepackage{float}
\usepackage{cite}
\usepackage[algo2e]{algorithm2e}

\usepackage{soul} 

 \definecolor{red}{rgb}{0.0, 0.0, 0.0}  
 \definecolor{blue}{rgb}{0.0, 0.0, 0.0} 

\renewcommand{\vec}[1]{\bm{#1}}
\DeclareMathOperator*{\argmin}{arg\,min}
\DeclareMathOperator{\sign}{sign}

\DeclareMathOperator{\Tr}{Tr}

\newcommand*{\trans}{^{\mkern-1.5mu\mathsf{T}}}
\newcommand*{\tran}{^{\mkern-1.5mu\mathsf{T}}}

\newcommand{\Wpm}{\vec{W}_{\pm}}
\newcommand{\Wzero}{\vec{W}_{0}}

\newcommand{\x}{{\vec{x}}}
\newcommand{\y}{{\vec{y}}}
\newcommand{\z}{{\mathbf{z}}}
\newcommand{\q}{{\vec{q}}}

\newcommand{\W}{{\vec{W}}}
\newcommand{\Wno}{{\vec{W}}}
\newcommand{\PP}{{\mathbf{P}}}

\newcommand{\Q}{{\mathcal{Q}}}
\newcommand{\J}{{\mathcal{J}}}
\newcommand{\R}{\mathcal{R}}
\newcommand{\xopt}{{\vec{x}^*}}
\newcommand{\A}{{\vec{A}}}
\newcommand{\s}{{\vec{s}}}

\newcommand{\Spm}{\vec{S}_{\pm}}
\newcommand{\Szero}{\vec{S}_{0}}

\newcommand{\E}{{\mathbf{E}}}
\newcommand{\F}{{\mathbf{F}}}

\newcommand{\proj}{{\vec{P}}}
\newcommand{\nullspace}[1]{{\mathcal{N}(#1)}}

\usepackage{lipsum}
\usepackage{amsfonts}
\usepackage{graphicx}
\usepackage{epstopdf}
\usepackage{algorithmic}
\ifpdf
  \DeclareGraphicsExtensions{.eps,.pdf,.png,.jpg}
\else
  \DeclareGraphicsExtensions{.eps}
\fi

\newsiamremark{remark}{Remark}
\newsiamremark{hypothesis}{Hypothesis}
\crefname{hypothesis}{Hypothesis}{Hypotheses}
\newsiamthm{claim}{Claim}

\headers{Bilevel learning of $\ell_{1}$ regularizers }{ Avrajit Ghosh, Michael T. McCann, Madeline Mitchell, Saiprasad Ravishankar}

\author{Avrajit Ghosh\thanks{A. Ghosh is with the Department of Computational Mathematics, Science and Engineering, Michigan State University, East Lansing, MI 48824, USA. Email: ghoshavr@msu.edu. *Corresponding author.} \and
Michael T. McCann\thanks{M. T. McCann is with the Theoretical Division, Los Alamos National Laboratory, Los Alamos, NM 87545, USA. Email: mccann@lanl.gov.} \and
Madeline Mitchell\thanks{M. Mitchell was a summer internee with the Department of Computational Mathematics, Science and Engineering, Michigan State University, East Lansing, MI 48824, USA. Email: mitch944@msu.edu.} \and
Saiprasad Ravishankar\thanks{S. Ravishankar is with the Department of Computational Mathematics, Science and Engineering and the Department of Biomedical Engineering, Michigan State University, East Lansing, MI 48824, USA. Email: ravisha3@msu.edu.}} 

\usepackage{amsopn}

\ifpdf
\hypersetup{
  pdftitle={Supervised Learning of Sparsity-Promoting Regularizers for Denoising},
  pdfauthor={}
}
\fi

\begin{document}
\title{Learning Sparsity-Promoting Regularizers using Bilevel Optimization}
\maketitle

\begin{abstract}
We present a gradient-based \textcolor{blue}{heuristic} method for supervised learning
of sparsity-promoting regularizers for denoising signals and images.
Sparsity-promoting regularization
is a key ingredient in solving modern signal reconstruction problems;
however, the operators underlying these regularizers  are usually either
designed by hand
or
learned from data in an unsupervised way.
The recent success of supervised learning 
(e.g., with convolutional neural networks) in solving image reconstruction problems
suggests that it could be a fruitful approach to designing regularizers.
Towards this end,
we propose to denoise signals using a variational formulation with a parametric, sparsity-promoting regularizer, 
where the parameters of the regularizer are learned to minimize the mean squared error of reconstructions on a training set of ground truth image and measurement pairs.
Training involves solving a challenging bilevel optimization problem;
 we derive an expression for the gradient of the training loss using the closed-form solution of the denoising problem and
provide an accompanying gradient descent algorithm to minimize it.
Our experiments with structured 1D signals and natural images 
\textcolor{blue}{indicate} that the proposed method can learn an operator that
outperforms well-known regularizers
(total variation, DCT-sparsity, and unsupervised dictionary learning)
and collaborative filtering for denoising.
\end{abstract}

\begin{keywords}
Sparse representations, denoising, analysis operator learning, transform learning, bilevel optimization, machine learning.
\end{keywords}

\begin{AMS} 
  68W65, 68U65.
\end{AMS}

\section{Introduction}

Sparse representations have been widely used in image processing and imaging.
In this paper, we develop an approach for supervised learning of nonsmooth sparsity promoting regularizers for image restoration.
We first review several categories of image restoration or reconstruction methods from the literature, before summarizing our contributions.
\subsection{Image Reconstruction}

Image reconstruction deals with recovering an image from its noisy measurements, which has applications in medicine (X-ray CT \cite{elbakri2002statistical,kim2014combining}, MRI \cite{vlaardingerbroek2013magnetic,larkman2007parallel,fessler2020optimization,ravishankar2010mr}, PET\cite{alessio2006pet,reader2007advances}, SPECT \cite{lyra2011filtering}), remote sensing \cite{campbell2011introduction,schowengerdt2006remote}, astronomical imaging \cite{akiyama2019first,levis2021inference}, etc. Image reconstruction falls under the broad category of inverse problems, which deal with recovering the underlying information from incomplete observations or measurements. A signal $\x$ can undergo some process to form measurements $\y$. Let the process be modeled as an operator $\A(\cdot)$, i.e., $\y = \A(\x) +\boldsymbol{\epsilon} $, with $\boldsymbol{\epsilon}$ being measurement noise. Then, given $\y$ and knowledge of what the process $\A(\cdot)$ is, we are interested in recovering $\x$.
In many applications, this inverse operation is ill-posed, i.e., there can exist infinitely many solutions $\x$ that are consistent with the forward operator and the measurements. Hence, we need some prior knowledge about $\x$ so that we choose meaningful solutions from the set of solutions that are consistent in some sense with the forward model. To phrase it in a formal way, we want to regularize the recovery problem in such a way that the solutions that fit our prior knowledge of $\x$ would be preferred over the solutions which do not fit the prior knowledge. Herein, comes the importance of the term regularization. 
 
\textcolor{red}{To preserve the consistency of the measurements with the physical forward model, one often minimizes the data consistency term $ \| \A (\x) - \y \|_2^2$. Moreover, we design a functional $ \R(\x)$, such that minimizing it would restrict the solution space based on our prior knowledge of the data.}
Thus, the effective objective is
$\mathcal{J}(\x) = \| \A (\x) - \y \|_2^2 + \R(\x)$, where $\x \in \mathbb{R}^n$ is the underlying image to be recovered,
$\y \in \mathbb{R}^m$ are its noisy measurements, 
$\A:  \mathbb{R}^n \to \mathbb{R}^m  $ is a measurement (forward) operator,
and $\R: \mathbb{R}^n \to \mathbb{R}$
is a regularization functional. So we solve the optimization problem:
\begin{equation}
   \hat{\x} = \underset{\x} \arg\min  \frac{1}{2} \| \A (\x) - \y \|_2^2 + \R(\x)
   \label{main}
\end{equation}
  
\subsection{Types of Regularizers}

Characterizing $\R(\x)$ to obtain a good reconstructor is a critical step in image reconstruction. 

An important regularization model is sparsity, popularized by advances in compressed sensing~\cite{1614066}. Sparsity of a signal or an image $\x$ is the number of nonzeros in the signal (viewed relative to its dimension) obtained via the $\ell_0$ ``norm"\footnote{The $\ell_0$ functional is not really a norm because it is not homogeneous.} $||\x||_{0}$. Minimizing $\mathcal{J}(\x) = \| \A \x - \y \|_2^2$ subject to an $\ell_{0}$ constraint $||\x||_{0} \leq s$ is an NP-hard problem. 
The popular convex relaxation of this nonconvex problem involves relaxing the $\ell_{0}$ ``norm" to the $\ell_{1}$ norm, which has been proven to preserve sparsity. 

Typically reconstructed images are expected to have certain well-known properties like smoothness, piecewise-continuity, or even local or nonlocal structural similarity. 
Rather than assuming the image is directly sparse, it is a common approach to find a sparse solution in a transform domain, i.e., we minimize $||\Wno \x||_{1}$, where the $\Wno$ matrix projects the image to a domain, where we expect it to be sparse. 
For example, 
$\Wno$ could be the finite difference matrix (this case is also known as total variation (TV) minimization) for piecewise-constant signals, or if the image is relatively smooth, $\Wno$ can be the discrete cosine transform (DCT) or the discrete wavelet transform (DWT) matrix depending on the nature of smoothness. 

In all of the above methods, the transform matrix $\Wno$ is known before-hand and hence the form of the regularizer is known. There has been much interest in the past decade to adapt the regularizer to data \cite{ravishankar2016data,ravishankar2012learning,ravishankar2010mr,rubinstein_analysis_2013}.
The regularization parameters such as the sparsifying transform could be learned from a dataset of high-quality images or even directly from measurements~\cite{ravishankar2010mr} (during image reconstruction).
\textcolor{red}{These methods are a type of unsupervised learning \cite{aharon2006k,ravishankar2010mr,ravishankar2012learning,8844696}.
Supervised learning \cite{7949028,zhang2018ista,10.5555/3157096.3157098,9747096,9363511,6737048,romano2017little} uses paired training data (typically ground truth images and their corrupted measurements) to learn the regularizer.}

\subsection{Contributions}
Learning sparsifying transforms in an unsupervised manner has been explored in several works.
In recent times, supervised methods in image reconstruction have gained much attention. Methods like Plug and play Priors (PnP) \cite{venkatakrishnan2013plug} and Regularization by denoising (RED) \cite{romano2017little} have been increasingly popular because of their flexibility and enhanced performance for image reconstruction. Although researchers have focused a lot of attention on analyzing the convergence of PnP and RED \cite{liu2021recovery,chan2016plug,buzzard2018plug}, plugging a black-box denoiser into an iterative algorithm may not have significant mathematical interpretation. Therefore, we propose a novel way of learning a sparsifying transform in a supervised fashion using paired training data to function within a variational formulation for denoising.

Consider the process where a signal/image is contaminated by noise to form noisy measurements. Suppose there are $T$ pairs of such clean signals and measurements denoted by $(\x_{t},\y_{t})_{t=1}^T$. The measurements ($\y_{t}\in\mathbb{R}^n $) are related to the clean signal ($\x_{t}\in\mathbb{R}^n $) as $\y_{t} = \x_{t} + \boldsymbol{\epsilon} $ where $ \boldsymbol{\epsilon} \sim \mathcal{N}(\mathbf{0},\sigma^2 \mathbf{I})$. 
We propose learning a sparsifying transform transform $\Wno \in \mathbb{R}^{n \times n }$ for denoising using the $T$ sets of paired data $(\x_{t},\y_{t})_{t=1}^T$ via the following formulation:

\begin{subequations}
\label{eq:bilevel}
\begin{align}
    & \underset{\Wno}{\arg \min } \;                     \Q(\Wno) = \frac{1}{T} \sum_{t=1}^{T} \;  \frac{1}{2} \|\x_{t}^*(\Wno, \y_t) - \x_t \|_2^2  \label{eq:upper-level} \\
   &  \textrm{s.t.} \;\, \x_{t}^*(\Wno, \y_t) = \underset{\x}{\arg \min} \| \x - \y_t \|_{2}^2 + ||\Wno \x||_{1}. \label{eq:lower-level}
\end{align} 
\end{subequations}

The intuition for learning such a transform $\Wno$ would be: \textit{If (1) $\x_{t}^*(\Wno, \y_t)$ is the result of denoising of $\y_t$ using a transform $\Wno$, then (2) the result $\x_{t}^*(\Wno, \y_t)$ should be close to the ground-truth $\x_{t}$}. These two statements are formulated mathematically as the lower-level problem and the upper-level problem respectively, in the above bilevel optimization. Statement~\textit{(1)} is cast as a lower-level denoising problem to denoise $\y_{t}$ using transform $\Wno$. Statement \textit{(2)} makes sure that the result of the lower-level problem $\x_{t}^*(\Wno,\y_t)$ is close to the ground-truth signal $\x_{t}$ in an $\ell_{2}$ norm sense (this could be replaced in general with other quality metrics). 

We solve this bilevel optimization problem using our proposed algorithm which we term as "Bilevel Learning of $\ell_{1}$ regularizers with Closed-Form Gradients" dubbed BLORC. Before conveying 
our approach of solving the optimization problem, we give a brief review of the existing methods of solving such bilevel optimization problems (Section~\ref{section:bilevel_intro}). In doing so, we point out the drawbacks of the existing algorithms which attempt to learn regularizers in bilevel optimization of the form \eqref{eq:bilevel}. A recent work of ours in~\cite{9747201} presented a very brief overview of our proposed approach along with some preliminary results. We extend the work here in terms of theory, analysis, derivation, and detailed experiments. We considered the denoising problem in this work to derive and demonstrate our approach and we believe the methods could be extended to other inverse problems as well. We summarize our contributions as follows:

\begin{enumerate}
    \item We propose a simple \textcolor{blue}{heuristic} approach for \eqref{eq:bilevel} by replacing the lower-level problem \eqref{eq:lower-level} with a derived closed-form expression. 
    \item We provide a detailed analysis of the nature of a sign-pattern (arising from the closed-form expression), which allows us to obtain explicit gradient expressions in local neighborhoods of the transform, hence making gradient descent possible.
    \item We address the drawbacks of the existing methods for solving  \eqref{eq:bilevel} which are discussed in detail in Section \ref{drawbacks} and we discuss how our method overcomes these drawbacks. 
    \item We perform experiments on image denoising on some images from Urban-100 Dataset and compare with methods like BM3D and unsupervised sparsifying operator learning methods like Analysis K-SVD~\cite{rubinstein_analysis_2013}.
    \textcolor{red}{To the best of our knowledge, our image denoising experiments are the first that involves learning sparsifying transforms by solving bilevel optimization with a non-smooth lower level cost.}
\end{enumerate}

\textcolor{red}{We note here that we are applying gradient descent to a function that is not differentiable everywhere (piecewise differentiable); while this is potentially theoretically concerning, it is common in practice, e.g. \cite{lee2020correctness,fiege2018algorithmic,lecun1999object}. We discuss this in more details in Section \ref{sec:locan}.} \textcolor{blue}{Our experiments indicate that our method works well in practice. In this paper, we present a heuristic method and the drawn conclusions are mainly based on empirical observations.  }


\subsection{Organization}
The remainder of the paper is organized as follows. Section~\ref{section:bilevel_intro} briefly summarizes the well-known approaches to tackle a bilevel optimization problem and how our proposed approach BLORC performs when compared to these approaches. Section~\ref{section:methods} discusses in detail the methods we propose to solve the bilevel optimization problem, which form the bedrock of our algorithm presented in Section~\ref{section:algorithm}. Experiments performed on 1D signals and 2D images using the BLORC algorithm are presented in Section~\ref{section:experiment}. Section~\ref{section:noise-levels} discusses how BLORC performs under various underlying noise levels and under what circumstances its performance breaks down.
Section~\ref{section:derivations} presents the mathematical proofs and derivations of results
presented in the methods section.
Finally, in Section~\ref{sec:conclusions}, we conclude.

\section{Bilevel optimization in inverse problems}
\label{section:bilevel_intro}
\textcolor{red}{Image reconstruction has been studied extensively under the lens of bilevel optimization  \cite{de2017bilevel,kunisch_bilevel_2013}}. In this section, we provide a brief overview of some well-known methods for solving such an optimization problem. Without loss of generality, we rewrite the bilevel optimization problem presented in \eqref{eq:bilevel} in a more general form as follows: 
\begin{gather*}
     \hat{\boldsymbol{\beta}} = \underset{\boldsymbol{\beta}}{\arg \min } \,  \mathcal{L} (\boldsymbol{\beta},\x^*(\boldsymbol{\beta}) )   \\
    \textrm{s.t.} \,\; \x^*(\boldsymbol{\beta}) = \underset{\x}{\arg \min} \,  \mathcal{G} (\x ,\boldsymbol{\beta} ).
    \label{general}
\end{gather*}
It comprises two optimization problems. The upper level problem minimizes the function $\mathcal{L} (\boldsymbol{\beta},\x^*(\boldsymbol{\beta}))$. This loss function has an argument $\x^*(\boldsymbol{\beta})$ which comes from solving a lower level optimization problem by minimizing $\mathcal{G} (\x ,\boldsymbol{\beta} )$. 
Here, $\boldsymbol{\beta}$ denotes optimizable parameters of the lower-level problem. In the general form, we represent the optimization variable as the vector $\boldsymbol{\beta}$ instead of a matrix $\Wno$ unlike in \eqref{eq:bilevel}. It is straightforward to extend all the derivations when the learnable parameter is a matrix instead of a vector.\footnote{Note that for a matrix $\Wno$, \eqref{eq:gen_imp} can be rewritten as $ \nabla \mathcal{L}(\Wno) = \nabla_{\Wno} \mathcal{L} (\Wno,\x^* ) + {(\nabla_{\Wno}\x^*(\Wno))} [\nabla_{\x^{*}}  \mathcal{L} (\Wno,\x^*(\Wno) )]$ \label{implicitmatrix}. In this case, $ \nabla \mathcal{L}(\Wno)$,$\nabla_{\Wno} \mathcal{L} (\Wno,\x^* ) $ are matrices and $ (\nabla_{\Wno}\x^*(\Wno))$ is an order 3 tensor and the vector $\nabla_{\x^{*}}  \mathcal{L} (\Wno,\x^*(\Wno) ) $ is multiplied to a dimension of the tensor. Hence, the product of the order 3 tensor and the vector ${(\nabla_{\Wno}\x^*(\Wno))} [\nabla_{\x^{*}}  \mathcal{L} (\Wno,\x^*(\Wno) )] $ is a matrix. The product of the order 3 tensor ($\mathbf{T}$) and a vector $\x$ is a matrix and is defined as $\sum_{i} \mathbf{T}(i,j,k)\x(i) $.} We refer the readers to the review article by Crockett et al.~\cite{crockett2021bilevel}, which summarizes in more detail the existing literature on bilevel optimization for image reconstruction.

In bilevel optimization problems, the variable $\boldsymbol{\beta}$ connects both the upper level and the lower level problems. It is easy to see that if $\x^*(\boldsymbol{\beta})$ can be explicitly written as a function of $\boldsymbol{\beta}$, then the bilevel problem can be converted into a typical single level optimization problem, which can be easily tackled by optimization techniques using the chain rule of differentiation. But this is rarely the case. For learning sparsifying transforms in a bilevel framework, the upper-level cost $\mathcal{L} $ is smooth and convex w.r.t. the reconstructions but the lower-level cost $\mathcal{G}$ is convex but non-smooth and non-differentiable. For these problems, $\x^*(\boldsymbol{\beta})$ can't be written explicitly in terms of $\boldsymbol{\beta}$.
    Assuming the function $\mathcal{G} (\x ,\boldsymbol{\beta} )$ is a strictly convex function of $\x$ (hence the lower level optimization problem has a unique minimizer), then using chain rule, we have

\begin{align}
    \nabla \mathcal{L}(\boldsymbol{\beta}) = \nabla_{\boldsymbol{\beta}} \mathcal{L} (\boldsymbol{\beta},\x^* ) + \underbrace{(\nabla_{\boldsymbol{\beta}}\x^*(\boldsymbol{\beta}))^T}_{\text{Implicit Gradient}} \nabla_{\x^{*}}  \mathcal{L} (\boldsymbol{\beta},\x^*(\boldsymbol{\beta}) ).
    \label{eq:gen_imp}
\end{align}

In our problem, where $\mathcal{G}$ is nonsmooth, we do not have an explicit expression for $\x^*(\boldsymbol{\beta})$ in terms of $\boldsymbol{\beta}$. Hence obtaining the term $ \nabla_{\boldsymbol{\beta}}\x^*(\boldsymbol{\beta})$ (also termed as implicit gradient) is the main challenge in solving such bilevel problems.  The methods to derive $\nabla_{\boldsymbol{\beta}}\x^*(\boldsymbol{\beta})$ can be broadly classified in four categories as briefly discussed in Sections \ref{IFT}, \ref{KKT}, \ref{Unrolling}, and \ref{Diffcp} . The derivation of the implicit gradient ($\nabla_{\boldsymbol{\beta}}\x^*(\boldsymbol{\beta}) $) using these methods follows closely with~\cite{crockett2021bilevel}. 

\subsection{Implicit Function Theorem (IFT)} \label{IFT}
In IFT, we assume that the lower level solution ($\x^*(\boldsymbol{\beta})$) can be expressed as an implicit function. Also, we assume that the lower level is unconstrained. Then there must exist a minimizer of $\mathcal{G}$ such that the gradient with respect to lower level optimization variable $\x$ at the minimizer $\x^*$ is zero, i.e., $\nabla_{\x} \mathcal{G}(\x^*(\boldsymbol{\beta}) ,\boldsymbol{\beta}) = \mathbf{0}$. \textcolor{red}{Using the chain rule to differentiate both sides,we can obtain an expression for $\nabla_{\boldsymbol{\beta}}\x^*(\boldsymbol{\beta})$ that when substituted in \eqref{eq:gen_imp} yields:}
\begin{align}
    \nabla \mathcal{L}(\boldsymbol{\beta}) = \nabla_{\boldsymbol{\beta}} \mathcal{L} (\boldsymbol{\beta},\x^*(\boldsymbol{\beta}) ) -  \nabla_{x \boldsymbol{\beta}} \mathcal{G}( \x^*,\boldsymbol{\beta})^T [\nabla_{xx}\mathcal{G}(\x^*,\boldsymbol{\beta})]^{-1}. \nabla_{\x^*}  \mathcal{L} (\boldsymbol{\beta},\x^*(\boldsymbol{\beta}) ).
    \label{eq:imp_exp}
\end{align}

\subsection{Using KKT conditions} \label{KKT}

\textcolor{red}{Replacing the lower level problem by  $\nabla_{x}\mathcal{G}(\x^{*}(\boldsymbol{\beta}),\boldsymbol{\beta}) = \mathbf{0}$, a similar expression (Equation \eqref{eq:imp_exp}) for calculating the gradient of the loss with respect to  $\boldsymbol{\beta}$ can be found by the method of forming a Lagrangian }
\textcolor{red}{
\begin{align*}
    \mathbf{L}(\x,\boldsymbol{\beta},\boldsymbol\gamma) = \mathcal{L} (\beta,\x)  + \boldsymbol\gamma^T \nabla_{x}\mathcal{G}(\x,\boldsymbol{\beta}),
\end{align*}
}
 \textcolor{red}{and finding an expression for $ \boldsymbol\gamma $ using the KKT stationarity condition $ \nabla_{x} \mathbf{L}(\x,\boldsymbol{\beta},\boldsymbol\gamma) =  \mathbf{0}$.} 
In problems that involve learning sparsifying transforms, the lower-level cost function $\mathcal{G}$ is non-smooth and non-differentiable w.r.t. the parameters $\boldsymbol{\beta}$. Hence in all the above formulations, $\nabla_{\boldsymbol{\beta}}\mathcal{G}(\x^*,\boldsymbol{\beta})$ is not-well defined over the domain of $\boldsymbol{\beta}$ rendering all the above expressions for $\nabla_{\boldsymbol{\beta}}\x^*(\boldsymbol{\beta})$ not well-defined in that whole domain. 

One way to avoid such a problem is to make a smooth approximation of the function that involves $\boldsymbol{\beta}$ (in this paper $\Wno$). One of the first works in this direction is of~\cite{peyre_learning_2011}. Here the bilevel problem was optimized for the case when the lower level problem had an analysis sparsity prior $||\Wno \x||_{1}$. A smooth approximation of the $\ell_{1}$ regularizer was made to make the lower level cost function smooth w.r.t. the dictionary coefficients. Then the parameters of the dictionary were learned using the minimizer approach using KKT conditions as discussed above. 

Another work~\cite{yunjin2012learning} learns both the $\ell_{1}$ norm-based analysis and synthesis operators using bilevel optimization with implicit differentiation.  However, the non-smooth $\ell_{1}$ penalty in the lower level problem was replaced by differentiable penalty functions to employ implicit differentiation.
In~\cite{mairal_task_2012}, the authors address a synthesis version of the bilevel problem,
wherein the reconstruction problem involves finding sparse codes, $\z$ such that $\|\x - \Wno\z \|$ is small.
This change from the analysis to synthesis formulation means that the optimization techniques used in~\cite{mairal_task_2012} do not apply here. In \cite{sprechmann_supervised_2013}, the authors derived gradients for a generalization of \eqref{eq:main} by relaxing $\|\Wno\x\|_1$ to $\min_{\z} \alpha\|\Wno\x - \z \|_2^2 + \|\z\|_1$.
This approach gives the gradient in the limit of $\alpha \to \infty$,
however the expression requires computing the eigendecomposition of a large matrix.
Therefore the authors use the relaxed version $\alpha < \infty$ in practice.
A brief preprint work~\cite{chen_learning_2014} derived a gradient for the upper level cost $Q(\Wno)$ in equation~\eqref{eq:bilevel}
by expressing $\Wno$ in a fixed basis and 
using a differentiable relaxation.
Finally,~\cite{chen_insights_2014}
provides a nice overview of the topic of analysis operator learning in its various forms,
and also tackles the bilevel optimization problem in \eqref{eq:bilevel} using a differentiable sparsity penalty. \textit{A drawback for all the above discussed works is approximating the non-smooth $\ell_{1}$ penalty with a smooth and differentiable penalty.} 

\subsection{Unrolling and related approaches}
\label{Unrolling}
The idea of unrolling approaches is to replace the lower level optimization problem by a sequence of iterations. Iterative algorithms have been used extensively to solve convex optimization problems. A very common optimization problem prevalent in image reconstruction is the unconstrained $\ell_{1}$ minimization problem which consists of an $\ell_{1}$ penalty and a data-fidelity term. ISTA (Iterative Shrinkage Thresholding Algorithm) algorithm and ADMM (Alternating direction method of multipliers) are well known iterative algorithms to solve this type of problem. Say if the algorithm used to solve the lower level problem is unrolled for $T$ iterations, then for each iterative update, let $\Psi$ denote the update as a function, i.e.,
\begin{align}
    \x^{(t)} = \Psi(\x^{(t-1)},\boldsymbol{\beta}) \;\; \textrm{for t = 1,2,...,T}. 
    \label{fixedpoint}
\end{align}
Assuming the algorithm runs for $T$ iterations, we will use  $\x^{(T)}$ as the approximate reconstruction in the upper level loss. 
Let $\mathbf{H_{t}} = \nabla_{x}\Psi(\x^{(t-1)},\boldsymbol{\beta})$ and $\mathbf{J_{t}} = \nabla_{\boldsymbol{\beta}}\Psi(\x^{(t-1)},\boldsymbol{\beta})$, then using chain rule yields
\begin{align}
    \nabla_{\boldsymbol{\beta}}\x^*(\boldsymbol{\beta}) \approx \mathbf{J_{T}} +  \sum_{t=1}^{T-1} (\mathbf{H}_{T}\mathbf{H}_{T-1}..\mathbf{H}_{t+1}) \mathbf{J_{t}}.
    \label{unroll_imp}
\end{align}

In the unrolling based approaches,  a chain of \textcolor{red}{sub-gradients} are obtained for the unrolled iterations. For example, \cite{ochs_bilevel_2015} differentiates the iterations of a non-linear primal primal-dual algorithm. Depending on the sequence order of calculating the sum in the implicit gradient \eqref{unroll_imp}, unrolling methods can be classified into forward and reverse mode. For the more common reverse mode~\cite{crockett2021bilevel}, the gradient computation starts from the $T$th step to calculate the product of $\mathbf{H}_{t}$'s. In each iteration, $ \mathbf{H}_{t+1}$ and  $ \mathbf{J}_{t}$ are calculated. In doing so, for each iteration, the corresponding $\x^{(t)} \in \mathbb{R}^{n \times 1}$ is stored in memory and the matrices $\mathbf{H}_{t+1}$ and  $ \mathbf{J}_{t}$ are calculated and stored in memory. 

\textit{Hence, the memory cost for the unrolling methods using the reverse mode becomes $\mathcal{O}(Tn +n^3)$. Also, unrolling for a large number of iterations may suffer from the problem of vanishing gradients.}
 
As the reconstruction may not be accurate if the algorithm solving the lower-level problem is unrolled only for finite iterations, researchers have recently proposed Deep Equilibrium models (DEM) \cite{gilton2021deep}. These models correspond to potentially infinite iterations of unrolling in the lower-level and hence may achieve better accuracy. DEM assumes that equation \eqref{fixedpoint} has a fixed point $\x^{\infty}$ such that a deep network or mapping $f$  with parameters $\theta$ can represent the fixed point as $ \x^{\infty} = f_{\theta}(\x^{\infty})$. The network parameters are learned to minimize the loss $L = \|\x_{t}- \x^{\infty}\|_{2}^2$ by an implicit differentiation of the fixed point equation $ \x^{\infty} = f_{\theta}(\x^{\infty})$. This helps convert the memory intensive task of backpropagating through many iterations of $f_{\theta}(.) $ to calculating an inverse Jacobian product, which can be further calculated by finding another approximate fixed point (solved by Neumann series). Hence, the task of finding the gradient of the loss $L$, i.e., $\frac{\partial L}{\partial \theta}$ just boils down to three major steps each iteration: a) calculating the residual, b) solving a fixed point equation by Neumann series (assuming it converges), and c)  multiplying gradient of network output $\frac{\partial f_{\theta}(\x^{\infty})}{\partial \theta}$ with the fixed point solution obtained in (b). Similar to DEM, our approach BLORC 
can be thought of as
unrolling the lower-level optimization 
to infinite iterations. This is because in BLORC, the lower-level is replaced by an implicit closed-form expression, just like the fixed point expression $ \x^{\infty} = f_{\theta}(\x^{\infty})$ in DEM is implicit. The difference is that we solve the lower-level problem in its exact form  for learning the transform $\Wno$ rather than relying on the mapping $f_{\theta}$ to
represent generic fixed points.

\subsection{Differentiating through a transformed cone problem (Diffcp)\cite{diffcp2019}}  \label{Diffcp}
Every convex optimization problem including LASSO type problems (\ref{eq:lower-level}) can be converted to a cone program which deals with minimizing a linear function over the intersection of a subspace and convex cone. A solution map is defined to be a function mapping numerical data defining the problem to the primal or dual solution of the problem. For example, the lower-level problem can be written as an implicit solution map $\mathbf{G}$ of the form $\x^*(\Wno) = \mathbf{G}(\y,\boldsymbol{\beta},\Wno)$. The objective of differentiating through a cone problem is to calculate how the perturbations in the variable ($\Wno$) affect the solution $\x^*(\Wno)$, i.e., calculating  $\x^*(\Wno + \delta \Wno) = \mathbf{G}(\y,\boldsymbol{\beta},\Wno+ \delta \Wno)$. This is done by decomposing the function $\mathbf{G}$ into a composition of differentiable atomic functions and then representing the derivative of each atomic function as an abstract linear map~\cite{diffcp2019} by a method called Automatic Differentiation (AD). For a LASSO type problem, there exist points in the numerical data domain ($\Wno$) on which the problem is non-differentiable. On these points, AD libraries compute heuristic quantities instead of derivatives~\cite{diffcp2019}. 
The drawbacks of autodifferentiation methods over BLORC are:
\begin{enumerate}
    \item \textit{In AD approaches, the task is divided into a sequence of differentiable operations as computational graphs on which backpropagation is performed through chain rule \cite{margossian2019review}. This division into a sequence of operations  and calculating gradients for each node of the graph can utilize significant memory.\footnote{For AD approaches, the memory requirement is dynamic, depending on the implementation of the expression graphs. Memory optimizing strategies \cite{margossian2019review} like retaping, checkpoints, and region-based memory aim to increase memory efficiency.}. This can be avoided if there is an analytical explicit expression for gradients.} 
    \item Hand-coded analytical method of calculating gradient is known to be the most accurate method \cite{bischof1995automatic} as it calculates exact derivatives. Hand-coded gradient methods are also known to have less ``Gradient to Function Compute Time Ratio" than AD methods (see Figure 1 in \cite{bischof1995automatic}). To further validate this, we perform an experiment in Section~\ref{sec:compcvx} to compare BLORC with the Diffcp implementation in terms of accuracy and time complextiy.
\end{enumerate}

\subsection{Drawbacks addressed by BLORC}
\label{drawbacks}
The drawbacks of the methods discussed in Sections~\ref{IFT}, \ref{KKT}, ~\ref{Unrolling}, and \ref{Diffcp} are briefly summarised as follows:
\begin{enumerate}
    \item  Minimizer methods (KKT/IFT) to learn sparsifying transforms $\Wno$ require calculating and storing the implicit gradient  $\nabla_{\Wno}\x^{*}(\Wno)$ to evaluate the gradient of the loss $\nabla \mathcal{L}(\Wno)$ as in the expression in the footnote $\ref{implicitmatrix}$. Storing the implicit gradient is of order $\mathcal{O}(n^3)$. 
    \item \textcolor{red}{Replacing the $\ell_{1}$ penalty by a smooth and differentiable approximator 
    may result in corner-rounding. See Section \ref{sec:comp-penalty} for comparison of $\ell_{1}$ penalty with smooth Huber loss.}
    \item  Unrolling the iterations of an algorithm (e.g., ISTA) for solving the lower-level variational problem and then back-propagating gradients (w.r.t. $\Wno$) through the iteration steps can
    suffer from storing 
    quantities
    for every iteration. 
    \item Unrolling method approximates the solution of the lower level problem $\x^{*}$ using a finite number of iterations $T$, \textcolor{red}{hence calculating only an approximation of the implicit gradient $\nabla_{\Wno}\x^{*}(\Wno)$.} 
\end{enumerate}

The method we propose alleviates such drawbacks and we summarize this in Table~\ref{comp_table}.

\begin{table}[hbt]
\centering
\begin{tabular}{ |l|l|l|l|l| } 
 \hline
 \multicolumn{1}{|p{3cm}|}{\centering Methods}
& \multicolumn{1}{|p{2cm}|}{\centering Minimizer \\ using \\ KKT/IFT \cite{peyre_learning_2011,yunjin2012learning,mairal_task_2012,sprechmann_supervised_2013,chen_learning_2014}}
& \multicolumn{1}{|p{2cm}|}{\centering Unrolled \\ methods \cite{ochs_bilevel_2015}}
& \multicolumn{1}{|p{1.5cm}|}{BLORC (Ours)}\\\hline
 \hline
 \multicolumn{1}{|p{3cm}|}{\centering Memory}  & $\mathcal{O}(n^3)$  & $\mathcal{O}(T + n^3)$  & $\mathcal{O}(n^2)$ \\
 
 \hline 
\multicolumn{1}{|p{3cm}|}{\centering Gradient \\ Accuracy}  & $\times$  & $\times$ & $\checkmark$  \\ 
 \hline
\end{tabular}
\caption{Comparison of some existing methods with our proposed approach BLORC to solve the bilevel optimization problem in (\ref{eq:bilevel}). For minimizer methods, storing the 
implicit gradient has cost $\mathcal{O}(n^3)$. For BLORC, only the gradient matrices require storing, so the memory cost is $\mathcal{O}(n^2)$.
We compare BLORC and Diffcp in Section~\ref{sec:compcvx}.
} \label{comp_table}
\end{table}

\section{Methods}
\label{section:methods}
To tackle the bilevel optimization problem, we proceed with the following steps:
\begin{enumerate}
    \item An implicit closed-form expression of the lower-level problem is derived using KKT stationarity conditions. We will show both theoretically (in some special cases) and  experimentally that in local neighborhoods of the transform, we get an explicit closed-form expression. 
    \item We take the differential of the locally explicit closed-form expression, and using matrix algebra~\cite{minka_old_2000}, explicit gradient expressions of the upper-level loss w.r.t. the rows of the transform are obtained. This helps us to avoid explicitly calculating and storing the implicit gradient $\nabla_{\Wno}\x^{*}(\Wno) $. Hence for BLORC, the memory cost is  $\mathcal{O}(n^2)$ (\ref{comp_table}) which involves just storing the gradient of upper-level loss ($\nabla_{\Wno}Q$) each iteration.
    \item Using the analytical gradient expression, mini-batch gradient descent is performed. 
\end{enumerate}
 
It is important to note two things in the 
bilevel problem \eqref{eq:bilevel}: 
\begin{enumerate}[label=\alph*]
    \item  No constraints are imposed on the matrix $\Wno$ while learning. This lets us learn more general operators.
    \item  We let the algorithm learn the scaling of the regularization penalty and hence we do not require a separate scalar regularization strength to be learned. ($\beta ||\Wno \x||_{1} = ||(\beta \Wno) \x||_{1})$). Thus, the lower level problem does not contain a scalar regularization parameter.
\end{enumerate}

\subsection{Closed-form solution obtained by duality}

Consider the lower-level  functional 
\begin{equation}
     \label{eq:analysis}
       \J(\x, \Wno, \y) = \frac{1}{2}\| \x - \y \|_2^2 +  \| \Wno \x \|_1,
 \end{equation}
 with $\x,\y \in \mathbb{R}^n$
 and
 $\Wno \in \mathbb{R}^{k \times n}$.
It is strictly convex in $\x$
(because the $\ell_2$ norm term is strictly convex
and the $\ell_1$ norm term is convex)
and therefore has a unique global minimizer.
Thus we are justified in writing $\x^*(\Wno) =\argmin_\x \J(\x, \Wno, \y)$
without the possibility of
the minimizer not existing or being a nonsingleton. Note that although $\x^*$ depends on $\y$ and $\Wno$,
the $\y$-dependence is not relevant for this derivation
and
we will not  notate it explicitly.
Our key observation in deriving the closed-form expression of~\eqref{eq:analysis}  is that we need to know the sign pattern of  $\Wno\x^*(\Wno)$. So, the closed-form expression is an implicit equation where the reconstruction $\x^*(\Wno)$ is itself dependent on $\sign (\Wno \x^*)$ (where $[\sign(\z)]_i$ 
is defined to be -1 when $[\z]_i < 0$;
0 when $[\z]_i = 0$;
and 1 when $[\z]_i > 0$). 
Let $k_{=0}$ denote the set  \{$i$ $\in$ (1,2,3,..,k) \,:\, $(\Wno \x^{*})_{i} =0 $\} and $k_{\ne0}$ denotes \{$i$ $\in$ (1,2,3,..,k) \,:\, $(\Wno \x^{*})_{i} \ne 0 $  \}. 
Also, let us define $\Wzero$ as the matrix containing the rows of $\Wno$ that are indexed by the set $k_{=0}$ and $\Wpm$ contains rows of  $\Wno$ that are indexed by the set $k_{\ne0}$. Then we have the following theorem. A similar closed-form expression has been obtained in equation (33) in \cite{tibshirani2011solution} using a different method. 


\begin{theorem}[Closed-form expression for $\argmin_\x \J(\x, \Wno)$]
\label{closed-form}
Let the nonzero pattern $\s$ denote $\sign (\Wno \x^* )_{\Wno \x^* \ne 0}$ and let $\Wzero$, $\Wpm$ contain the rows of $\Wno$, whose indices are given by the sets $k_{=0}$ and $k_{\ne0}$, respectively.
Then the closed-form expression of the optimization problem $ \x^* = \arg \min_{\x} \frac{1}{2}\| \x - \y \|_2^2 +  \| \Wno \x \|_1$ is obtained from Lagrangian dual analysis as 
\begin{equation}  
   \x^*(\Wno)= \proj_\nullspace{\Wzero} (\y - \Wpm \tran \s ),
   \label{eq:closed_form}
\end{equation}
where $\proj_\nullspace{\Wzero} $ is the projector matrix onto the nullspace of $\Wzero$ and is given by  $\proj_\nullspace{\Wzero} = (\vec{I} - \Wzero^+ \Wzero)$.
\end{theorem}


To obtain the closed-form expression for a particular point on the domain of $\Wno$, we solve the lower-level problem \eqref{eq:lower-level} to obtain the reconstruction $\x^{*}(\Wno)$. This is because $\Wpm, \Wzero, \s$ in~\eqref{eq:closed_form} depend on $\x^{*}(\Wno)$. One may ask the question of why should we use an implicit closed-form equation if we can find the reconstruction by methods like ADMM or PGD? But we re-emphasize that our objective in this step is not to find $\x^*$ but to obtain a closed-form expression that allows us to take gradients with respect to $\Wno$.  
 
 \subsection{Local analysis of the closed-form expression}
 \label{sec:locan}
  Generalized LASSO-type problems do not have an explicit closed-form solution~\cite{tibshirani2011solution}. In fact, the closed-form expression we derived in~\eqref{eq:closed_form} is an implicit equation. Note that $\x^*(\Wno)$ is a function of the sign pattern $c(\Wno) = \sign (\Wno \x^*(\Wno))$ because the row-splits $\Wzero$ and $\Wpm$ are based on $c(\Wno)$. But again the sign pattern $c(\Wno) $ itself depends on $ \x^*(\Wno)$. The equation is implicit because $ \x^*(\Wno) $  appears on both sides; however when $ \x^*(\Wno) $ is restricted to lie in a certain region, the dependency on $ \x^*(\Wno) $  in the right hand side may be dropped, making the equation explicit. Thus, if the sign-pattern $c(\Wno) $ is constant in a neighborhood of $\Wno$, the equation (right-hand side) is explicit in the neighborhood. So, to take gradients of the expression in~\eqref{eq:closed_form} 
 w.r.t. 
 $\Wno$, it is important to establish the fact there exists an $\epsilon$-neighborhood 
 $[ \triangle \Wno \in \mathbb{R}^{k \times n} | \max(|  \triangle \Wno |_{ij}) \leq \epsilon]$ such that $c(\Wno +  \triangle \Wno) = c(\Wno)$. Only then, we can treat the equation~\eqref{eq:closed_form} as an explicit equation w.r.t. $\Wno$ in an $\epsilon$-neighborhood of $\Wno$. \textcolor{red}{For clarification, we do not assume this pattern is constant on the whole domain, but only in a neighbourhood and this allows us to derive gradient expressions.}
 Since there exists no explicit closed-form for $\x^*(\Wno)$, it is difficult to prove that the sign-pattern remains constant in an $\epsilon$-neighborhood in general. So, we analyze two cases when $\x^*(\Wno)$ has an explicit closed-form expression.
 
 \subsubsection{The scalar denoising problem}
Consider the scalar denoising problem
$x^*(w) = \argmin_x \frac{1}{2}(x-y)^2 + |w x|$ and $c(w) =\sign (wx^*(w))$. The measurement is assumed to be generated as $y = x_{t} + \epsilon$, where $x_{t}$ is the ground-truth.
Assuming that $y\ge0$,
one can show that $x^*(w) = y-|w|$ when $y-|w| \ge 0$ and $0$ otherwise.
As a result, $c((0, y)) = 1$, $c((-y, 0)) = -1$, and $c((-\infty, -y] \cup 0 \cup [y,\infty))=0$ is piecewise constant. A similar result holds when $y\le0$.
Thus $\Q(w)= (x^*(w) - x_{t})^{2}$ is smooth except at $w=0,-y,y$. \textcolor{red}{Note that $\Q(w)$ is differentiable in the intervals near the
minimizers of $\Q$ and the non-differentiable points form a set of zero measure.}
\begin{figure}[!t]
    \centering
     \includegraphics[width=0.75\textwidth]{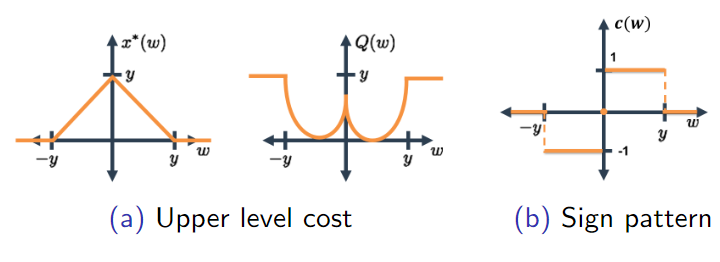}
    \caption{\textcolor{red}{The sign pattern $c(w) = \sign (wx^*(w))$ is piecewise constant for the scalar denoising problem. Also, the upper-level cost $Q(w)$ is piecewise smooth with respect to $w$. Note that its non-differentiable points are also not local/global minimizers.}}
    \label{fig:signs}
\end{figure}

\subsubsection{Denoising problem under orthogonal constraint}
\label{section:ortho}
Consider the denoising problem $\x_t^*(\Wno) = \underset{x}{\arg \min} \frac{1}{2}\| \x - \y_t \|_{2}^2 + \lambda||\Wno \x||_{1}$ but under an 
orthogonal constraint on (square)
$\Wno$ (lying on the Stiefel manifold), i.e., $\Wno \Wno^T  = \mathbf{I}$. It is easy to show that the solution to this optimization problem is given by $\x_t^*(\Wno) = \Wno^T \mathcal{S}_{\lambda}(\Wno \y_t)$ where $\mathcal{S}_{\lambda}(.)$ denotes the soft-thresholding function with parameter $\lambda$. Then the sign-vector $c_t(\Wno) = \sign(\Wno \Wno^T \mathcal{S}_{\lambda}(\Wno \y_t) ) = \sign(\mathcal{S}_{\lambda}(\Wno \y_t) )$. Considering each of the indices, $[c_t(\Wno)]_{i}$ is  piecewise constant except where $|(\Wno \y_t)_{i} |= \lambda $. Hence, $c_t(\Wno)$ is piecewise constant in each of the coordinates except at points where $ |(\Wno \y_t)_{i} |= \lambda$. We show in Section~\ref{section:Stiefel} that these points form a set of Lebesgue measure zero over the Stiefel manifold. Thus, the upper level cost function $Q(\Wno)$ can be seen to be a piecewise smooth function over orthogonal $\Wno$ matrices except on a set of measure zero.
 
 \subsubsection{Experimental validation for constant sign pattern}
 In the generalized setting for $\Wno$, the denoising problem does not have a closed-form expression. So, to experimentally check our claim, we perform an experiment consisting of the following steps:
 \begin{enumerate}
     \item Setting $\Wno = \mathbf{I}$, the lower-level denoising problem is solved using a noisy signal $\y$ to obtain  $\x^*(\Wno)$ and hence using that to obtain $c(\Wno) = \sign (\Wno \x^*(\Wno))$.
     \item Perturbing $\Wno$ in an $\ell_{\infty}$ radius ball, we obtain the sign-pattern in a
     neighborhood of $\Wno$ as $c(\Wno + \triangle \Wno) = \sign ((\Wno + \triangle \Wno) \x^*(\Wno + \triangle \Wno))$ such that $ \max(|  \triangle \Wno |_{ij}) \leq \eta$ for small $\eta>0$.
     \item  We then checked if $c(\Wno) = c(\Wno + \triangle \Wno)$. 
     Our experiments showed that there always exists a $\triangle \Wno$ with any $\eta = 10^{-8}$ (machine precision is $10^{-16}$) for which $c(\Wno) = c(\Wno + \triangle \Wno)$. For a particular point on the domain of $\Wno$, we performed thousands of trials of generating $\triangle \Wno$ with $\eta = 10^{-8}$, and everytime $c(\Wno) = c(\Wno + \triangle \Wno)$ was satisfied. We also performed the experiment with $\Wno$ being a 1D-finite difference matrix, a zero matrix,  and also a random matrix with its entries drawn from $\mathcal{N}(0,1)$ and also with different measurements ($\y$) and observed similar results. 
 \end{enumerate}
In the experiments above, we chose the measurement $\y$ to have length $64$ and the matrix dimensions were $64 \times 64$.

\subsection{Gradient calculations}
\label{Grad_cal}
Next, we compute the
gradient of the scalar $Q(\Wno)$ with respect to $\Wno$
using locally explicit closed-form expressions for the lower level problems.
Second, for the gradients to exist for a specific training signal's loss, the sign pattern vector $\sign (\Wno \x^* )$ has to remain constant in an open set containing $\Wno$. Only then the closed-form expression for $\x^*(\Wno)$ 
is valid in each region where $\sign (\Wno \x^* )$ is constant.

We refer the readers to
\cite{minka_old_2000}, which summarizes how to generate derivatives (Jacobian) of scalars with respect to matrices using differentials. We briefly summarize how we obtain gradients of the upper level scalar loss $Q(\Wno)$ with respect to sub-rows ($\Wzero,\Wpm$) of the transform $\Wno$ using differentials. 
 
We denote the Jacobian of the scalar $Q$ w.r.t. $\Wno$ 
as  $\nabla_{\Wno}Q$, which has the same dimensions as $\Wno$ as it is just $[\frac{\partial Q}{\partial w_{ij}}]$ where $w_{ij} $ is the element on the $i$th row and $j$th column of matrix $\Wno$. The differential of the scalar $\partial Q(\Wno)$ is defined as $\partial Q(\Wno) = Q(\Wno + \partial \Wno) -Q(\Wno)$, where $\partial \Wno$ is the differential of the matrix $\Wno$, i.e., $\partial \Wno = [\partial w_{ij}]_{i,j}$, and 
has the same dimension as $\Wno$. 
 
From~\cite{minka_old_2000}, we see that if we can write $ Q(\Wno + \partial \Wno)$ as $ Q(\Wno) + \Tr(\E^{T}\partial\Wno )$, then the Jacobian 
$\nabla_{\Wno}Q =\E $. Note that the canonical form $\Tr(\E^{T}\partial\Wno )$ is important to take the derivative of a scalar with respect to a matrix. So, a general rule of thumb to calculate such derivative of scalar w.r.t. matrix is
 \begin{enumerate}
    \item Compute the differential relating the scalar ($\partial Q$) and the matrix $\partial \Wno$. 
    \item Rewrite the result into the canonical form $ \partial Q(\Wno) = \Tr(\E^{T}\partial\Wno )$  to derive the Jacobian $\nabla_{\Wno}Q =\E$.\footnote{When we use the term differential $\partial Q$, $\partial \x^*(\Wno)$ or $\partial \Wno$ (same dimensions as $Q$, $\x^*(\Wno)$, and $\Wno$, respectively), we just mean the incremental changes in the scalar, vector, or matrix. 
A relationship such as $\partial Q = \Tr(\E^{T} \partial \Wno)$  indicates how the scalar $Q$ changes for incremental changes in $\Wno$. 
And when we use 
$\nabla_{\Wno}Q$, we mean the derivative of the scalar $Q$ w.r.t. $\Wno$. 
And the derivative can be obtained from the differential by writing it in the canonical form. That is if we have a differential relation $\partial Q(\Wno) = \Tr(\E^{T}\partial \Wno )$, then the derivative $\nabla_{\Wno}Q =\E$.}
 \end{enumerate}
 
For our problem, we don't have a direct relationship between $Q$ and $\Wno$ to get a differential relation like $\partial Q(\Wno) = \Tr(\E^{T} \partial \Wno )$. But rather the relation between $\partial Q$ and $\partial \Wno$ is connected by an intermediate differential variable $\partial \x^*(\Wno)$. $ \partial\x^*(\Wno)$ is the differential of the reconstruction in the lower-level problem which is a function of the transform $\Wno$. 
Since we have already defined a locally explicit expression for $\x^*(\Wno)$ in \eqref{eq:closed_form}, 
we can relate
$\partial \x^*(\Wno)$ and $\partial \Wno$ using \eqref{eq:closed_form} as discussed next. The proof for the following theorem is discussed in Section~\ref{section:derivations}.

\begin{theorem}[Differential of the closed form]
 Let $c(\Wno)$ denote the sign pattern of the reconstructed signal in the transform domain $\Wno$. Then if  $c(\Wno) = \sign (\Wno \x^*)$ is a constant vector in an open neighbourhood containing $\Wno$, then the gradient of the closed form in \eqref{eq:closed_form} w.r.t. $\Wno$ exists and the differential form is given by

   \begin{subequations}\label{eq:differntials}
     \begin{align}
     & \partial \xopt =  - \proj_\nullspace{\Wzero}  \partial \Wpm\trans \s \label{eq:differential1}\\ 
     &  \partial \xopt  = -(\Wzero^+ \partial\Wzero \proj_\nullspace{\Wzero}
    +  (\Wzero^+ \partial \Wzero \proj_\nullspace{\Wzero})\trans ) (\y - \beta \Wpm\trans \s), \label{eq:differential2} 
     \end{align}
    \end{subequations} 
where $\partial \xopt$, $\partial \Wzero$, and $\partial \Wpm$ are the differentials of $\xopt$, $\Wzero$, and $\Wpm$, respectively. 
\end{theorem}

Note that the closed-form expression in \eqref{eq:closed_form} for $\x^*(\Wno)$ is in terms of the row-submatrices. 
To obtain the sub-matrices $\Wpm$ and $\Wzero$ from $\Wno$, we need to have the corresponding sign vector. 
This is obtained by solving the lower level problem \eqref{eq:lower-level} 
using the current transform $\Wno$. 
Note that the differential expression \eqref{eq:differntials} is valid only in a neighborhood of the transform where the sign pattern $c(\Wno) = \sign (\Wno \x^*)$ is constant w.r.t $\Wno$.

To relate $\partial Q$ with $\partial \Wpm$ and $\partial \Wzero$, recall that 
$Q(\Wno) = \frac{1}{2} \|\x^*(\Wno) - \x_t \|_2^2$ (restricting for now to one term in the upper level loss). 
Hence the gradient of $Q$ w.r.t. $\x^{*}$ is $ \nabla_{\x^{*}}Q = (\x^*(\Wno) - \x_t) $. The canonical form for derivative of a scalar with respect to a vector is given as $\partial Q = \nabla_{\x^{*}}Q^{T} \partial \x^{*}$ \cite{minka_old_2000}. 
Using the expression for  $\nabla_{\x^{*}}Q$, we have $ \partial Q =(\x^*(\Wno) - \x_t)^{T}\partial \x^{*} $. Substituting for $\partial \x^{*}$ from \eqref{eq:differntials} and re-arranging into the canonical form, we can get two expressions of the form $ \partial Q = \Tr(\E^{T} \partial \Wpm)$ and $ \partial Q = \Tr(\F^{T} \partial \Wzero)$, from which the gradients w.r.t. $\Wpm$ and $\Wzero$ are straightforward. These are shown next and are derived in Section~\ref{grad_close}.
\begin{corollary}
\label{corollary}
Let $Q(\Wno) = \frac{1}{2} \|\x^*(\Wno, \y_t) - \x_t \|_2^2 $ be the upper level cost function that is smooth with respect to the intermediate reconstruction $\x^*(\Wno)$, then we derive an expression for the gradient of the cost $Q(\Wno)$ with respect to $\Wzero$ and $\Wpm$ as
\begin{align}
    \nabla_{\Wpm} Q &= -\s \nabla_\xopt Q \trans    \proj_\nullspace{\Wzero}
    \label{eq:grad1}\\
    \nabla_{\Wzero} Q &= -  (\proj_\nullspace{\Wzero} 
    (\vec{q} \nabla_\xopt Q \trans  + \nabla_\xopt Q \vec{q}\trans)
    \Wzero^{\dagger})\trans, 
    \label{eq:except}
\end{align}
with $\vec{q} = \y_{t} -  \Wpm\trans \s$.
Here,  $\nabla_\xopt Q = (\xopt - \x_{t})$.
\end{corollary}
It is important that the sign pattern $c(\Wno)$ (and $\x^{*} (\Wno)$ itself) is accurate for the analytical gradient expression to be correct. Hence, during gradient descent w.r.t $\Wno$, we first find $\x^{*}(\Wno)$ by running a large number of iterations of ADMM using the current transform $\Wno$. This step is the bottleneck of our proposed algorithm. 

\section{BLORC Algorithm}
\label{section:algorithm}
We perform minibatch gradient descent to learn the transform matrix $\Wno$ in a supervised manner from training pairs $(\x_{t},\y_{t})$. 
In all our experiments, we start from $\Wno_{0} = \vec{I}_{n \times n}$, the identity matrix. 
At the start of each epoch during training, the training pairs are randomly shuffled to remove dataset bias. Each sample in a batch is processed as follows.

\SetKwInput{KwInput}{Input}                
\SetKwInput{KwOutput}{Output}   
\SetKwInput{Kwinitialize}{Intialization}
\SetKwInput{Kwpreprocess}{Pre-processing}

\begin{algorithm}[H]\label{alg:1}
\KwInput{ $M$ pairs of training signals $(x_{t},y_{t})_{t=1}^M$, each of dimension $n \times 1 $, initial transform matrix $\Wno_{0} =I_{n \times n}$,
batch size $B$, number of epochs $E$, learning rate $\alpha$, threshold $\gamma$. 
}
\KwOutput{Trained $\hat{W}$}
\Kwpreprocess{For inputs that are clean and noisy images of size $N \times N$, extract patches of size $\sqrt{n} \times \sqrt{n}$ with stride $r$ and vectorize them.}
\Kwinitialize{$\Wno = \Wno_{0}$.}

 \For {\textbf{each epoch} }{
 Shuffle training pairs $(\x_{t},\y_{t})_{t=1}^M$. \\
 a) \For {\textbf{each training index in batch } } {
 
 \textbf{1) Obtain sign vector:}\\
 ${\mathbf{\x^*} = \underset{\mathbf{\x}}{\arg \min }  \| \y_{t} - \mathbf{\x} \|_{2}^2 +  \lambda\| \Wno\x \|_{1}}$ /*Solve iterations of ADMM to obtain $\x^*$ \\
 $\mathbf{\s}  = \sign (\mathbf{W} \mathbf{\x^*})_{|[\mathbf{W} \mathbf{\x^*}]_{i}| \geq \gamma}$  /* Obtain the sign vector  \\
 \textbf{2)  Pre-computations for gradient:}\\
 $\nabla_{\x^*} Q = 2(\x^* - \x_{t}) $ , $\mathbf{q} = \y_{t}  -  \Wpm\trans \mathbf{\s}$ \\
 Split rows of  $\Wno$ into $\Wpm$ and $\Wzero$ based on $\sign (\mathbf{W} \mathbf{\x^*}) $
 
 \textbf{3) Gradient Calculation: }\\
 $\nabla_{\Wpm} Q = -\mathbf{\s} (\nabla_{\x^*} Q)^T \proj_\nullspace{\Wzero}  $ /* Obtain gradient for $\Wpm$\\
  $\nabla_{\Wzero} Q = -  (\proj_\nullspace{\Wzero} 
    (\vec{q} \nabla_\xopt Q \trans  + \nabla_\xopt Q \vec{q}\trans)
    \Wzero^+)\trans$   /* Obtain gradient for $\Wzero$\\
\textbf{4) Stack the gradient matrices based on row partition}\\
$\nabla_{\Wno}Q = [\nabla_{\Wzero}Q;\nabla_{\Wpm}Q] $\\
\textbf{5) Accumulate the gradients $\nabla_{\Wno}Q$ for the current batch.}\\
}
b) Update $\Wno$ after each batch is processed. \\
$\Wno = \Wno - \alpha \nabla_{\Wno} Q$}
$\hat{\Wno} = \Wno $\\
\textbf{Post-processing} If training on image patches, then reshape rows of $\Wno$ to display convolutional filters.
\caption{BLORC}
\end{algorithm} 

\begin{enumerate}
    \item  Given the measurements $\y_{t}$ and current $\Wno$, the lower-level reconstruction problem is solved iteratively using ADMM to obtain an estimate of $\x_{t}^*(\Wno)$. 
    \item The sign vector $\sign (\Wno \x_{t}^*(\Wno))$ is obtained after hard-thresholding $\Wno \x_{t}^*(\Wno)$ with a small threshold parameter $\gamma$. The sign vector is of paramount importance as it decides the row-split of $\Wno$ into $\Wzero$ and $\Wpm$. \textcolor{red}{Hence, we run ADMM for enough iterations to obtain empirical convergence of the sign pattern.} 
    \item Then, using \eqref{eq:grad1} and \eqref{eq:except}, we 
obtain the
gradient of the upper level cost on a single training pair, i.e.,  $\nabla_{\Wno} Q$  which is the row-concatenation of  $\nabla_{\Wzero} Q$  and  $\nabla_{\Wpm} Q$. Averaging the gradients over the samples in the batch yields the minibatch gradient.   
\end{enumerate}

At the end of each batch, we update the matrix $\Wno$ based on the learning rate $\alpha$ and the mini-batch gradient. The updated $\Wno$  is used in the next batch in step (1) above.

Our method of gradient calculation can also be extended when the training pairs are image patches instead of 1D signals. For image denoising experiments, image patches of size $\sqrt{n} \times \sqrt{n}$ with an overlap stride of $r$ were extracted from images. Then the 2D patches are converted to 1D arrays as a pre-processing (first) step. In the end, the rows of the learned $\Wno$ are reshaped to look like convolutional filter patches. Except this first and last step, all the intermediate steps are the same for both 1D signals and image patches.
 
\section{Experimental results}
\label{section:experiment}
To demonstrate how well BLORC learns the transform matrix $\Wno$, we perform a series of denoising experiments for 1D signals and also for images from the Urban-100 dataset.

\subsection{Denoising 1D signals}
Here, synthetic 1D signals were generated to be sparse with respect to a specific transform that also provides a baseline to compare our learned transforms with.

\begin{figure}[!t]
    \centering
     \includegraphics[width=0.9\textwidth]{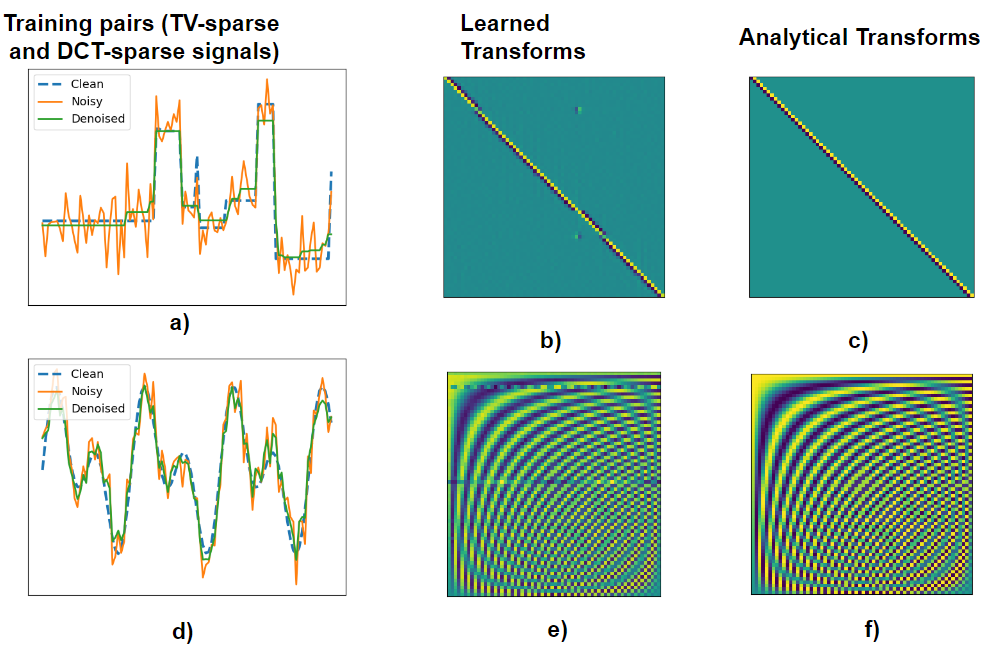}
    \caption{1D training pairs ($\x_{t}, \y_{t}$) (left column) and their corresponding learned transform $\hat{\Wno}$ (middle column).
    \textbf{a)}  TV-sparse training signal and its denoised version using learned transform. The noisy signal was denoised using the learned transform. \textbf{b)} Learned transform for piecewise constant signals. \textbf{c)} The analytical finite difference matrix (TV). \textbf{d)} DCT-sparse training signal and its denoised version using learned transform. \textbf{e)} Learned transform for DCT-sparse signals. \textbf{f)} The analytical 1D DCT matrix }
    \label{fig:lertran}
\end{figure}

In our experiment, we generate $M= 4000$ training pairs $(\x_t,\y_t)$, where the $\x_t$'s are piece-wise constant signals of length $n=64$ (with peak value normalized to 1)  and $\y_t$'s are noisy versions with additive i.i.d. Gaussian noise with standard deviation $\sigma = 0.1$. Figure~\ref{fig:lertran}(a) shows a single pair of such $(\x_t,\y_t)$. We perform minibatch gradient descent with batch size $B=100$ and run the algorithm for $E=750$ epochs. The learning rate was chosen to be $\alpha = 10^{-4}$ and the sign threshold was $\gamma = 10^{-3}$. The learned transform is shown in Figure~\ref{fig:lertran}(b). We repeated the experiment with the same parameters but with 
$x_t$'s chosen as
smoothly varying signals of different harmonics
that are sparse in the discrete cosine transform (DCT) domain as in Figure~\ref{fig:lertran}(d).
The learned transform is row-rearranged such that it has maximum correlation with the 1D-DCT matrix and is shown in Figure~\ref{fig:lertran}~(e). The learned transforms of Figures~\ref{fig:lertran}(b) and \ref{fig:lertran}(e)  capture the same intuition as the standard finite difference transform and the 1D-DCT transform, respectively, but in addition, they also have slight novel features 
learned for the denoising task. Experimental results \textcolor{blue}{indicate} that the learned transforms perform better than the standard transforms for denoising on a test-set. The average PSNR for 20 piecewise-constant test signals denoised using the learned $\Wno$ was 26.2 dB whereas that using the standard transform was 25.8 dB, with the PSNR of the noisy signals being 19.5 dB.\textcolor{red}{ The regularization parameter for the standard transform was chosen using golden-section search so that the reconstruction minimized the upper-level cost. }

We also perform a 1D denoising experiment comparing an unsupervised analysis dictionary learning method and our supervised approach BLORC.

For the unsupervised approach, the analysis dictionary ($\Wno$) is learned by minimizing the objective  $\sum_{t=1}^T \|\Wno \x_{t}\|_1$, where $\x_{t}$ is the $t^{th}$ clean signal, but enforcing an orthogonal constraint on $\Wno$ to avoid trivial zero solutions. We minimize the unsupervised training objective with $T=100$ piecewise constant training samples using the ADMM algorithm
with split variables $\z_t = \Wno \x_{t}$, and where orthogonality is enforced by solving an orthogonal Procrustes problem using 
a singular value decomposition (SVD) each iteration \cite{schonemann1966generalized}. 
\begin{figure}
    \centering
    \includegraphics[width=\textwidth]{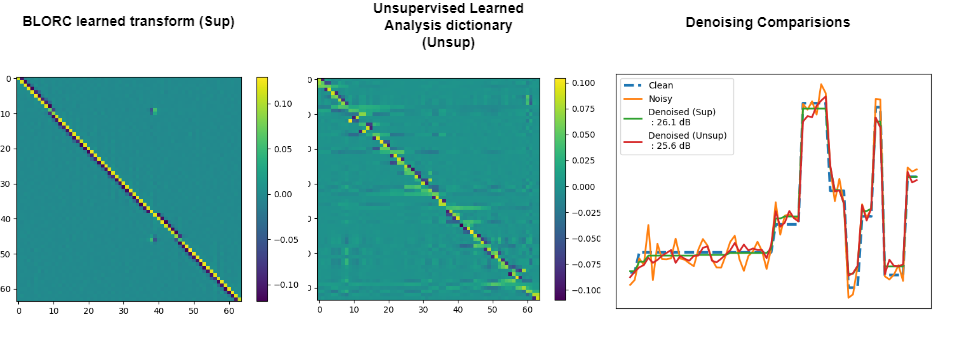}
    \caption{Learned transforms in BLORC and with unsupervised analysis dictionary
learning and denoising comparisons.}
    \label{fig:unsup}
\end{figure}

Denoising was performed on a test set of 15  piecewise-constant test signals (different from the training set) using ADMM. The average PSNR using the BLORC learned transform was 26.2 dB and that with the unsupervised learned analysis dictionary was 25.2 dB. The learned transforms along with example denoised signals are shown in Figure~\ref{fig:unsup}, where we observe that the rows of the BLORC learned matrix capture the "jumps" of the piecewise constant signals more efficiently.

\subsection{Image denoising}
Extending the BLORC algorithm to image patches, we learn reasonable transforms as well. We first demonstrate a simple experiment to denoise toy images 
which have directional stripes. This is a simple verification experiment, because we can guess how the sparsifying convolutional filters for such images may look.

We chose images (normalized) of size $256 \times 256$ with directional patterns (vertical stripes and diagonal stripes) and generated their noisy versions with i.i.d. Gaussian noise with $\sigma =0.1$. We extracted image patches of 
size $8 \times 8$ with an overlap stride of $7$. The reason for choosing a large stride was to ensure that distinct training pairs are obtained. As a preprocessing step, the image patches of size $8 \times 8$ were vectorized to 
size $n=64$.

\begin{wrapfigure}{r}{0.5\textwidth}
\vspace{-0.2in}
  \begin{center}
    \includegraphics[width=0.5\textwidth]{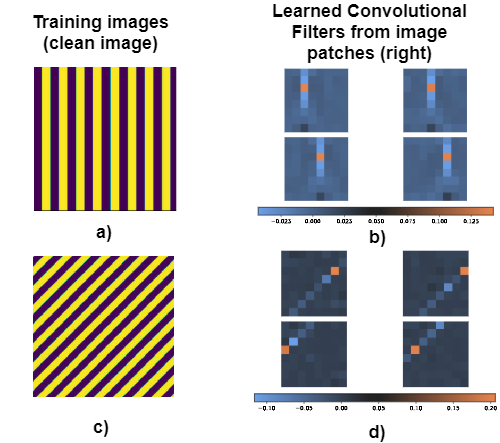}
  \end{center}
  \caption{Training images (only clean) with a specific orientation (left) and learned sparsifying filters (right): \textbf{a)} image with vertical stripes (vectorized patches extracted from the image  are used as training set), \textbf{b)} learned filters for vertical stripes, \textbf{c)} image with diagonal stripes, and \textbf{d)}  learned filters for diagonal stripes.}
  \label{fig:imagestrips}
\end{wrapfigure}

All the parameters and the intermediate steps of the minibatch gradient descent were unchanged from the 1D signal experiment above. The learned convolutional filters in  Figs.~\ref{fig:imagestrips}(b) and \ref{fig:imagestrips}(d) look orthogonal to their image counterparts in Figs.~\ref{fig:imagestrips}(a) and  \ref{fig:imagestrips}(c), respectively, which is the expected output. 
\textit{In particular, these transforms are learned not only to sparsify the images but to minimize the gap between reconstructions and the ground truth.}

Next, we show simple illustrative image denoising results using the Urban-100 dataset. The study is meant to provide some understanding of the behavior of supervised learning via BLORC and not an exhaustive demonstration of image denoising performance.
 We compare the results of image denoising by BLORC with the methods BM3D~\cite{dabov2006image} and Analysis KSVD~\cite{rubinstein_analysis_2013}. Both these algorithms are well-known to work on images with blocks, structures and orientations and hence comparing with these denoising algorithms gives us a standard baseline for the performance of BLORC. We fine-tuned the hyper-parameters of these algorithms to achieve the best denoising PSNR at test time. 

\begin{figure}[!t]
    \centering
    \includegraphics[width=\textwidth]{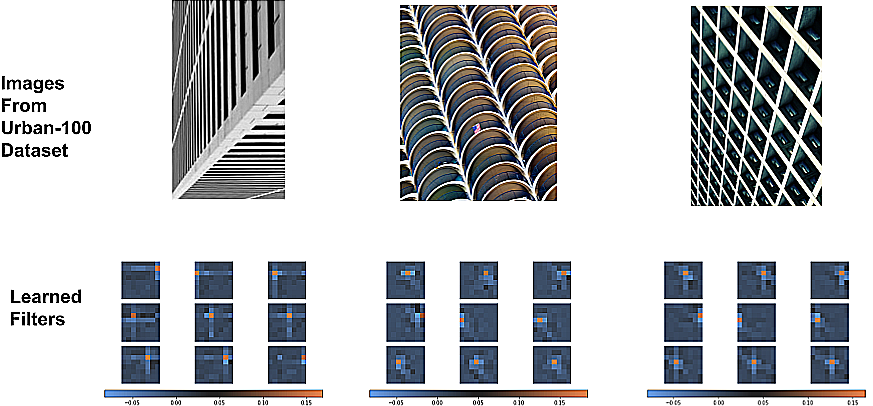}
    \caption{Patch-based Filters learned from images in the Urban-100 dataset. The filters capture orientations in the images. For each image, we show a subset of the rows of the learned $\Wno$ as patches/filters. }
    \label{fig:urban100}
\end{figure}

For image denoising with BLORC, the images are split into training and validation sets. In Figure~\ref{fig:urban100}, we show the three images in the training set used to learn three sets of convolutional filters. We denote these learned transforms by $\Wno_{1},\, \Wno_{2}$, and $\Wno_{3}$, respectively. The dimension of each of these learned transforms is $n \times n$, with $n$ being the unravelled patch dimension. In the experiments, we chose the square patch dimension to be $8\times8$, hence $n=64 $. Note that these filters have been learned at a patch level, i.e., by extracting patches from the training images.  We have purposefully chosen the training set to contain images with different orientations so that we learn convolutional filters capturing different orientations in a supervised manner. To use the learned filters for denoising images in a test set, we formed a row-stacked version of these three trasnsforms which we denote as $\Wno_{learned} \in \mathbb{R}^{3n \times n} $.

To test the effectiveness of the learned filters for denoising, we use this stacked transform matrix $\Wno_{learned}$ to denoise a different set of test images, labeled Image-1 to Image-4, respectively as shown in Table~\ref{Tab:image_Table}. 
We solve the denoising problem at the patch level, where we solve the following optimization problem using ADMM to obtain denoised patches  $\hat{\x}_{j}$:
\begin{gather*}
    \left \{ \hat{\x}_{j} \right \}
 =  \underset{ \left \{ \x_{j} \right \} }{\arg \min }  \sum_{j=1}^{n} (\| \x_{j} - \PP_{j} \y \|_{2}^2 + \| \Wno_{learned} \x_{j}   \|_{1} )
 \end{gather*}
The operator $\PP_{j}$ extracts the $j$th patch from the image. 
The regularizer makes sure that the solution at the patch level is sparse w.r.t. the transform $\Wno_{learned}$.
The denoised image is obtained after spatially aggregating the denoised patches ${\hat{\x}_{j}}$. 

\begin{table}[htbp]
\begin{center}
\begin{tabular}{|l|c|c|c|c|} \hline
Image/method & Clean Image & BM3D &    Analysis KSVD & BLORC (Ours) \\ \hline
 Image-1 &  
\includegraphics[width = 2.8cm]{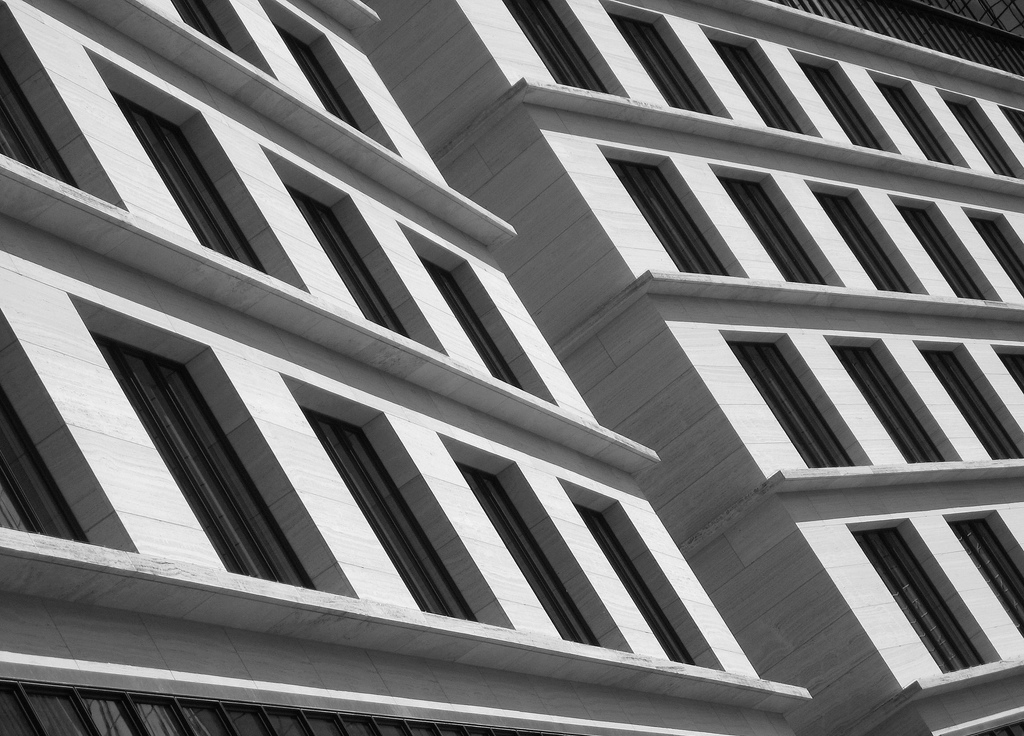}   & \includegraphics[width = 2.8cm]{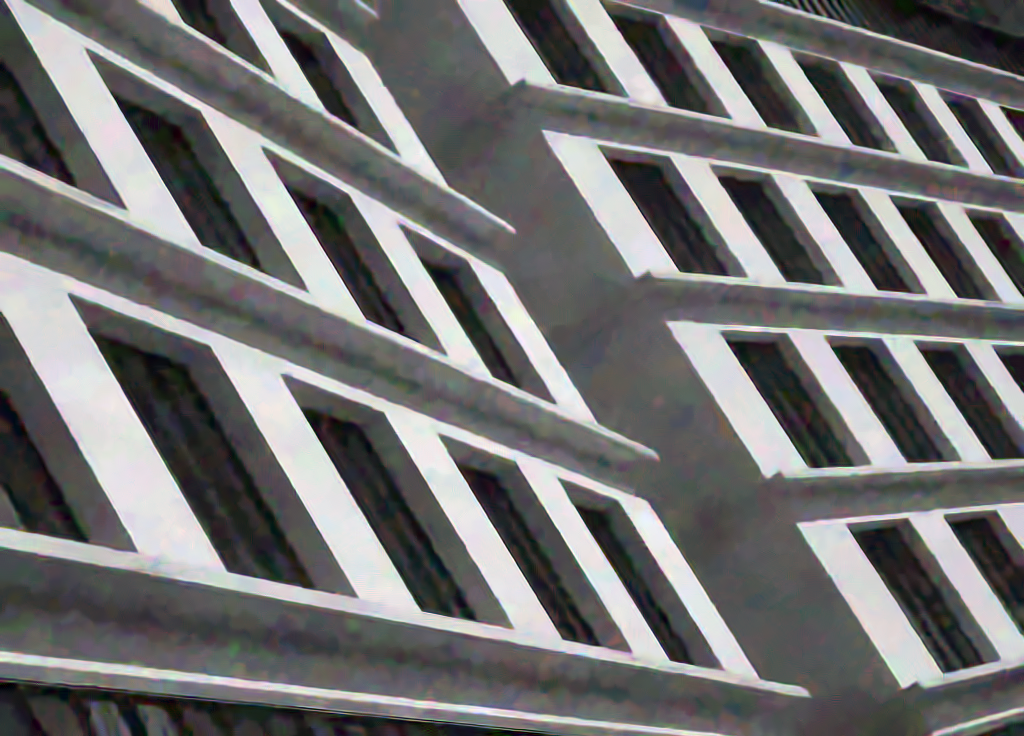}   & \includegraphics[width = 2.8cm]{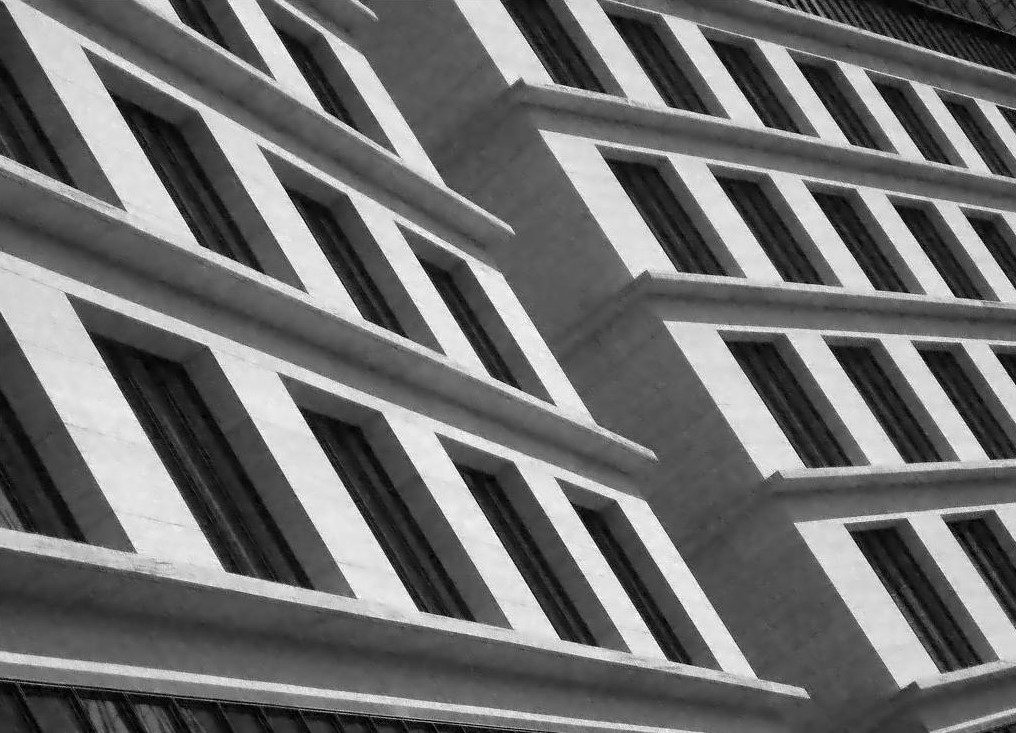} & \includegraphics[width = 2.8 cm]{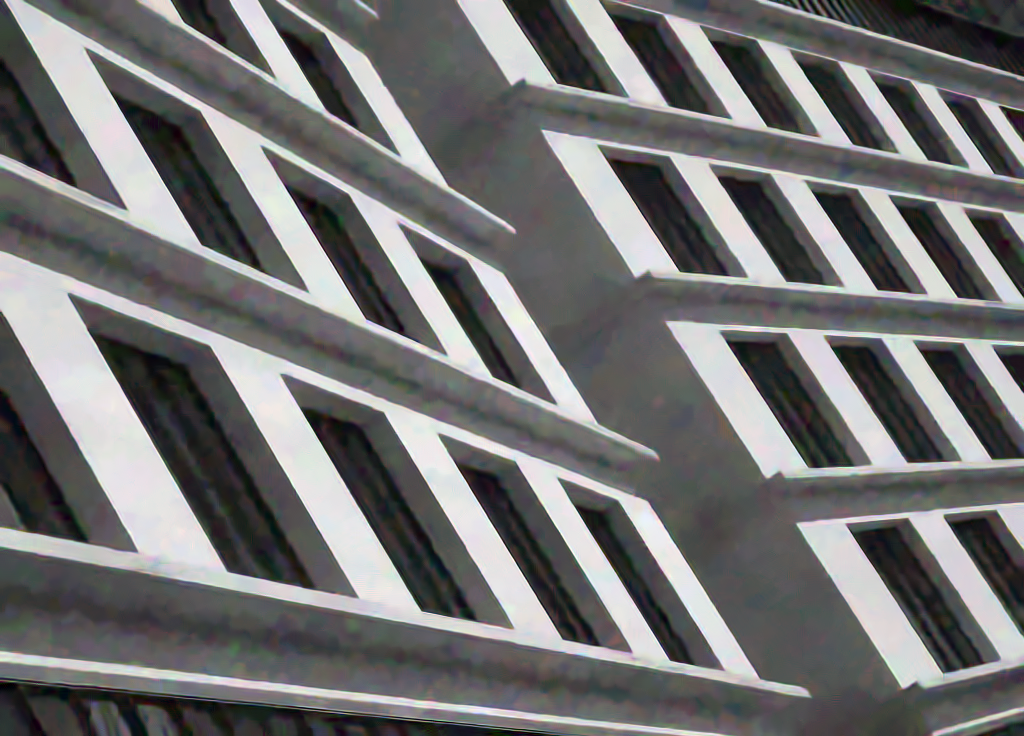} \\
\hline 
PSNR &  - &  26.27     & 25.92  & 27.02 \\
\hline
Image-2 &  
 \includegraphics[width = 2.8 cm]{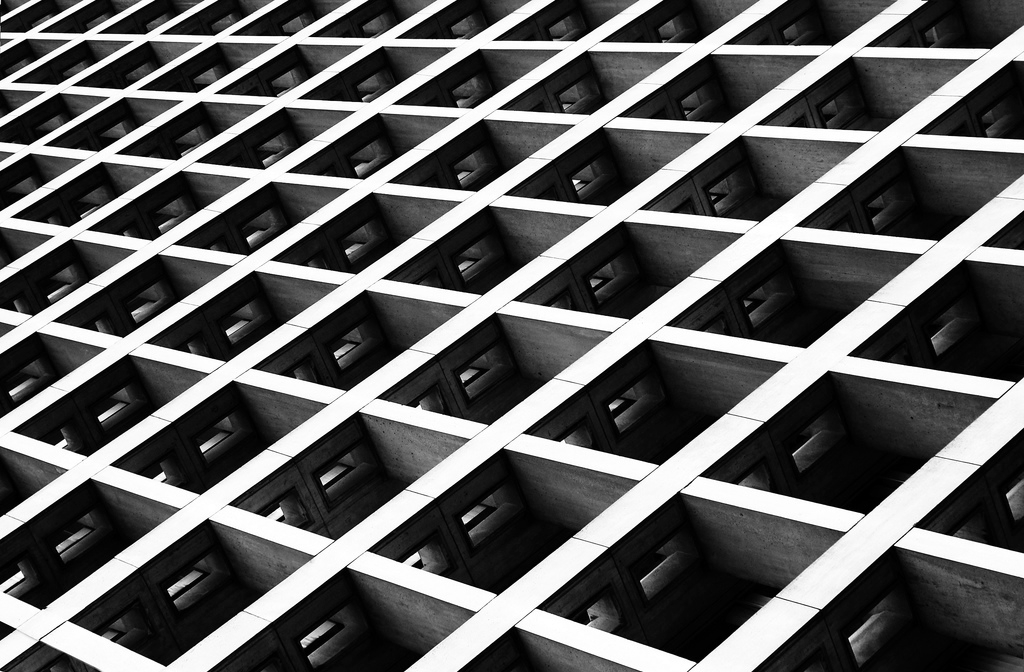}   & \includegraphics[width = 2.8 cm]{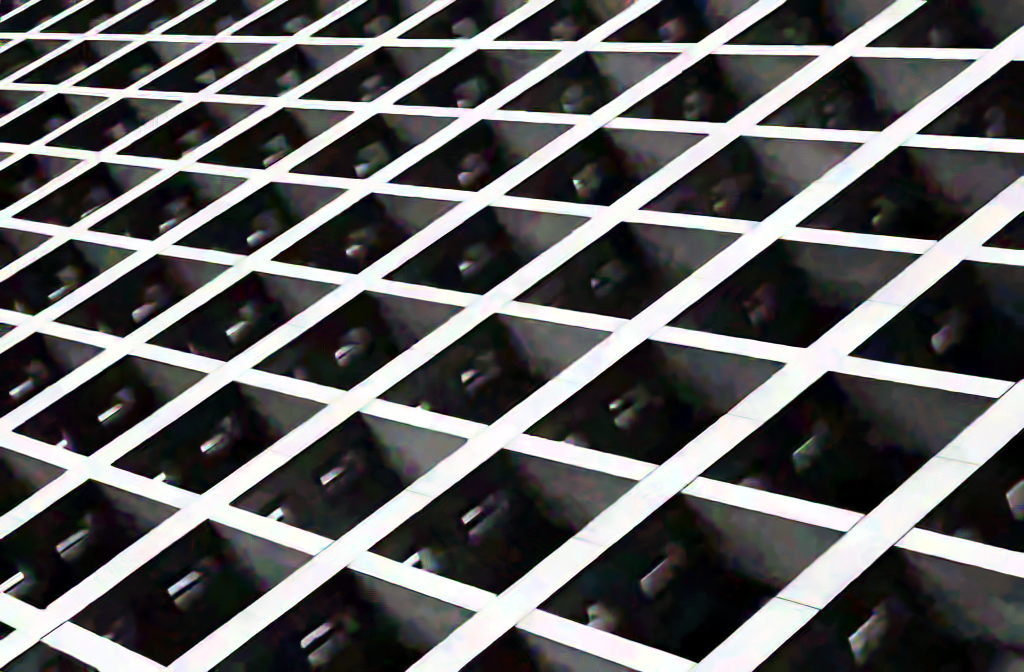}   & \includegraphics[width = 2.8 cm]{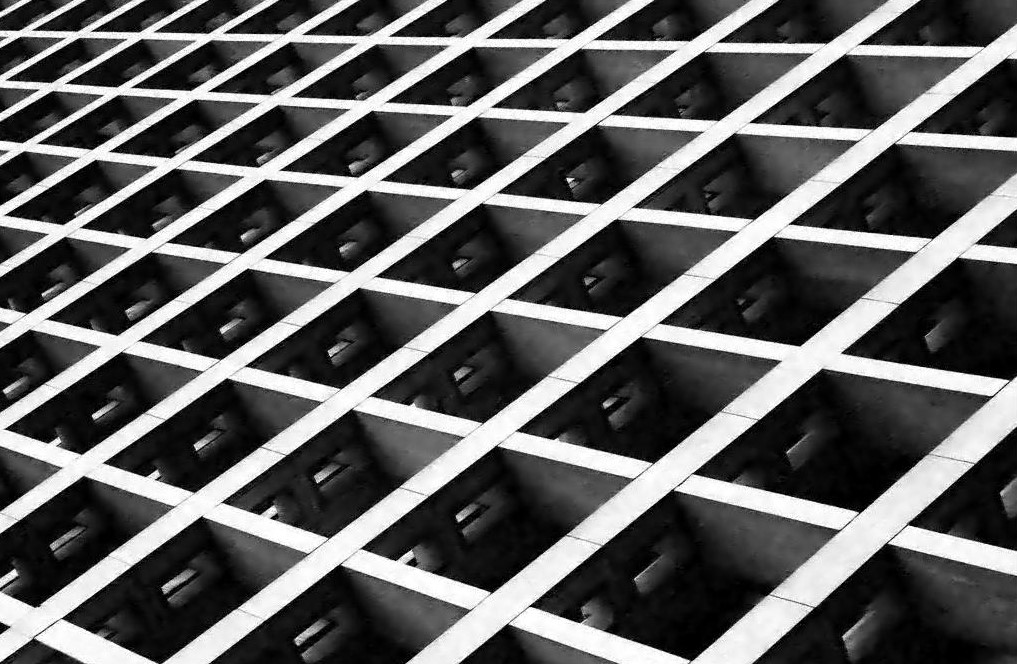} & \includegraphics[width = 2.8 cm]{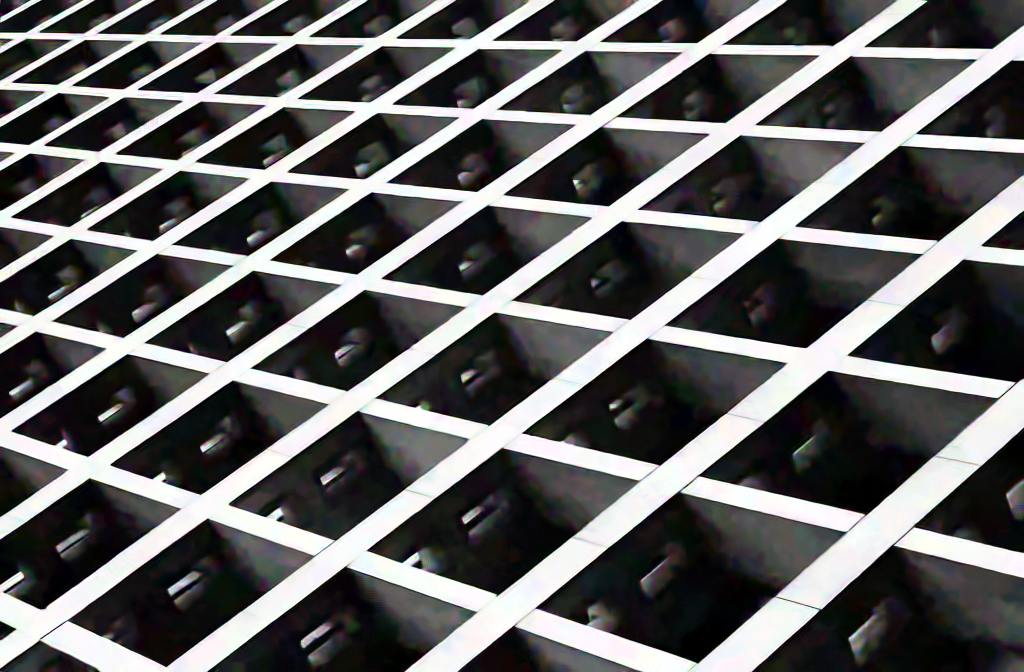} \\
\hline
PSNR &   -  &   25.51  & 25.30 & 25.20 \\ \hline \hline
Image-3 &  
\includegraphics[width = 2.8cm]{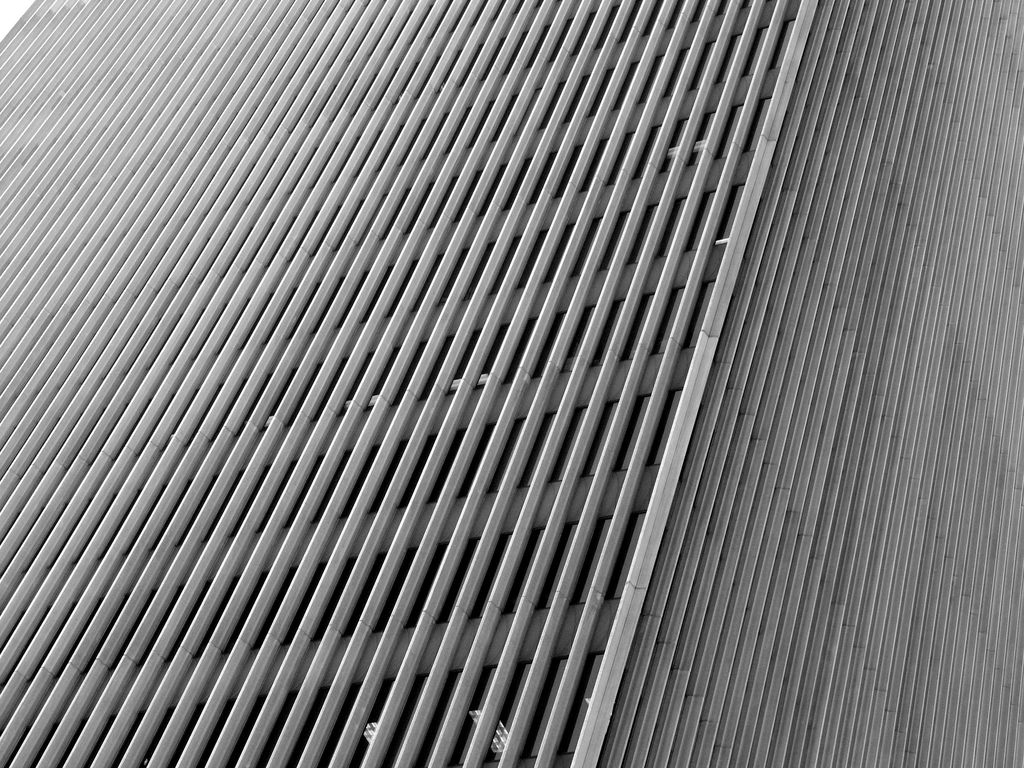}  & \includegraphics[width = 2.8cm]{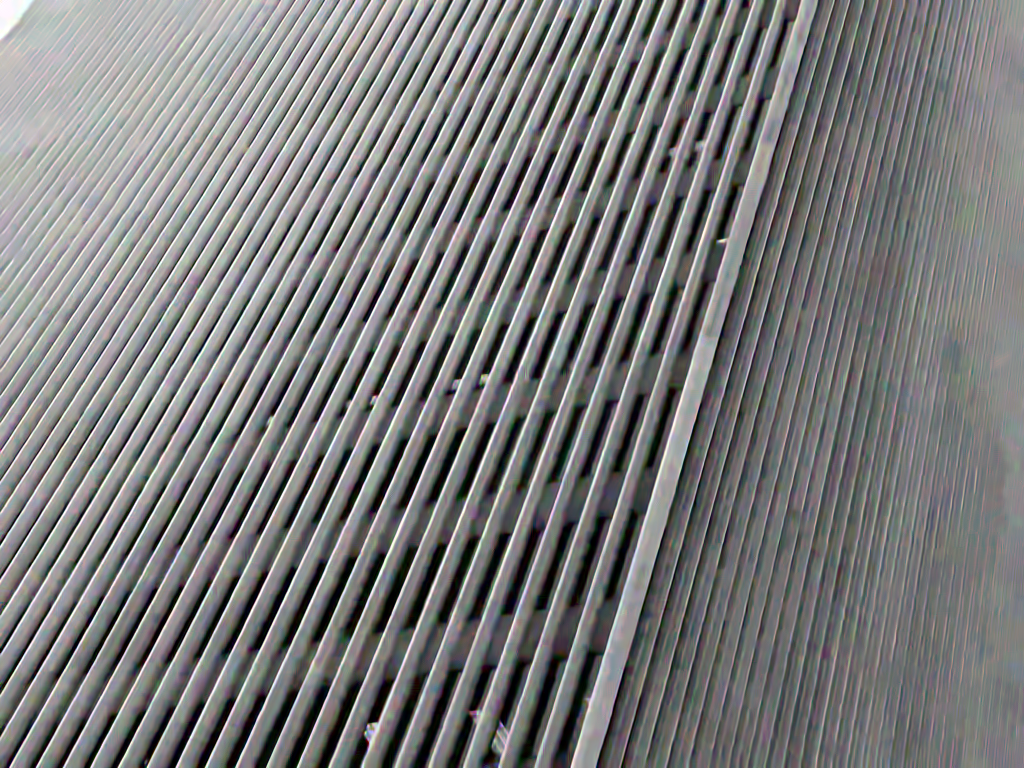}  & \includegraphics[width = 2.8cm]{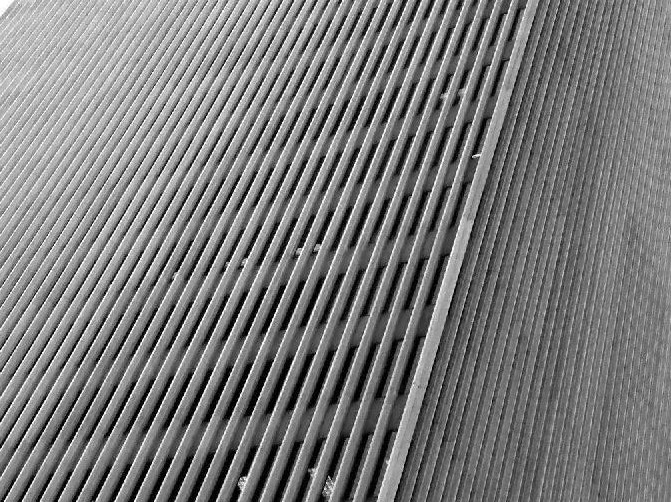} & \includegraphics[width = 2.8cm]{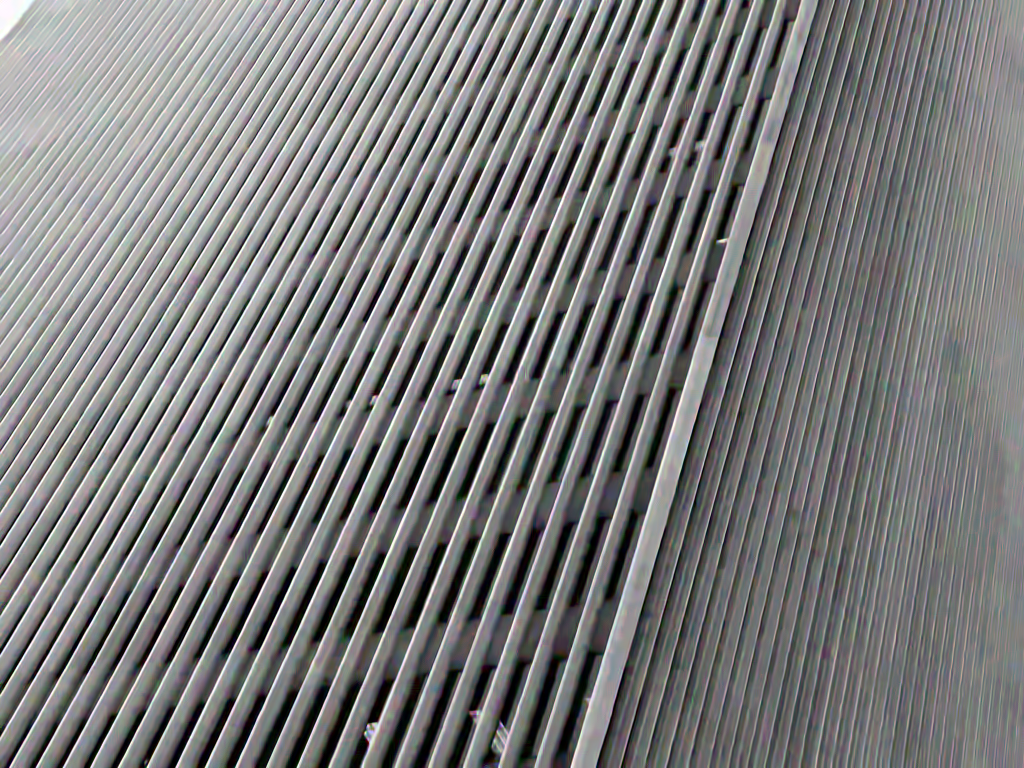} \\
\hline
PSNR &  -     &    21.90 & 21.34 & 22.17 \\ \hline \hline
Image-4 &  
\includegraphics[width = 2.8cm]{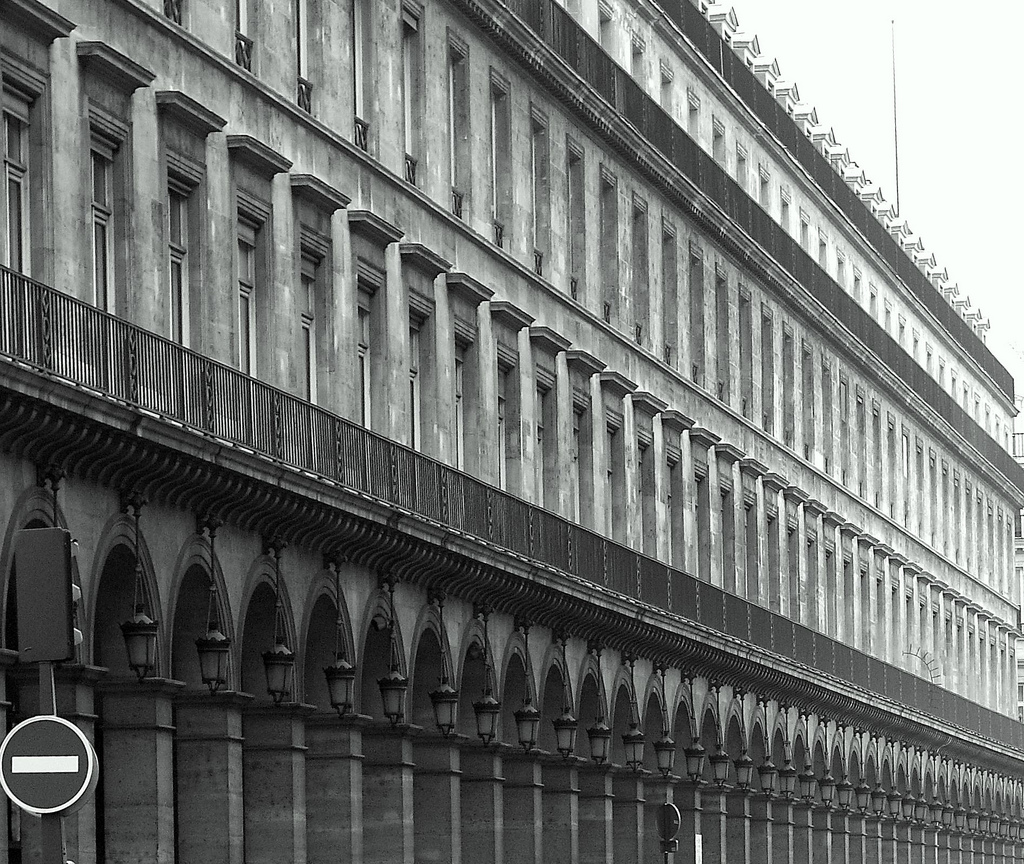}   & \includegraphics[width = 2.8cm]{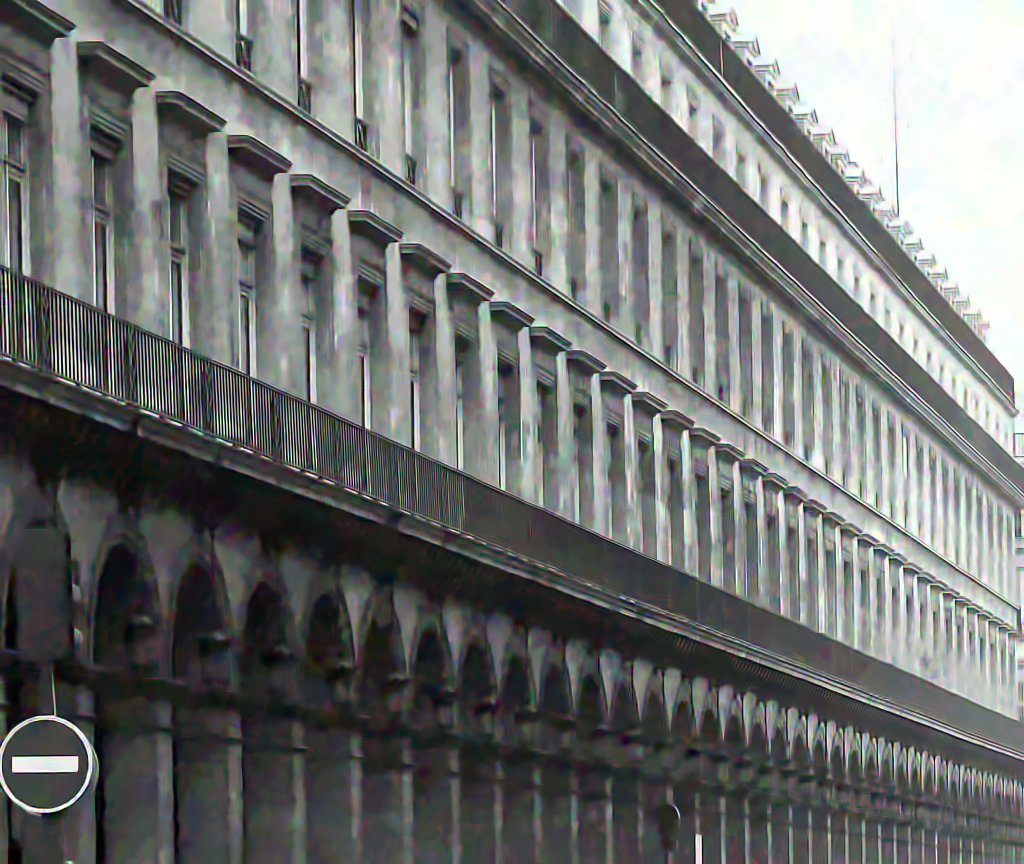}   & \includegraphics[width = 2.8cm]{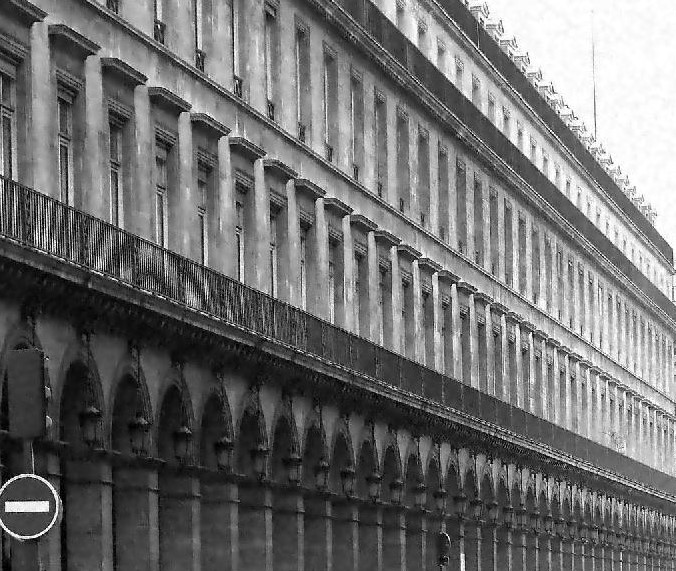} & \includegraphics[width = 2.8cm]{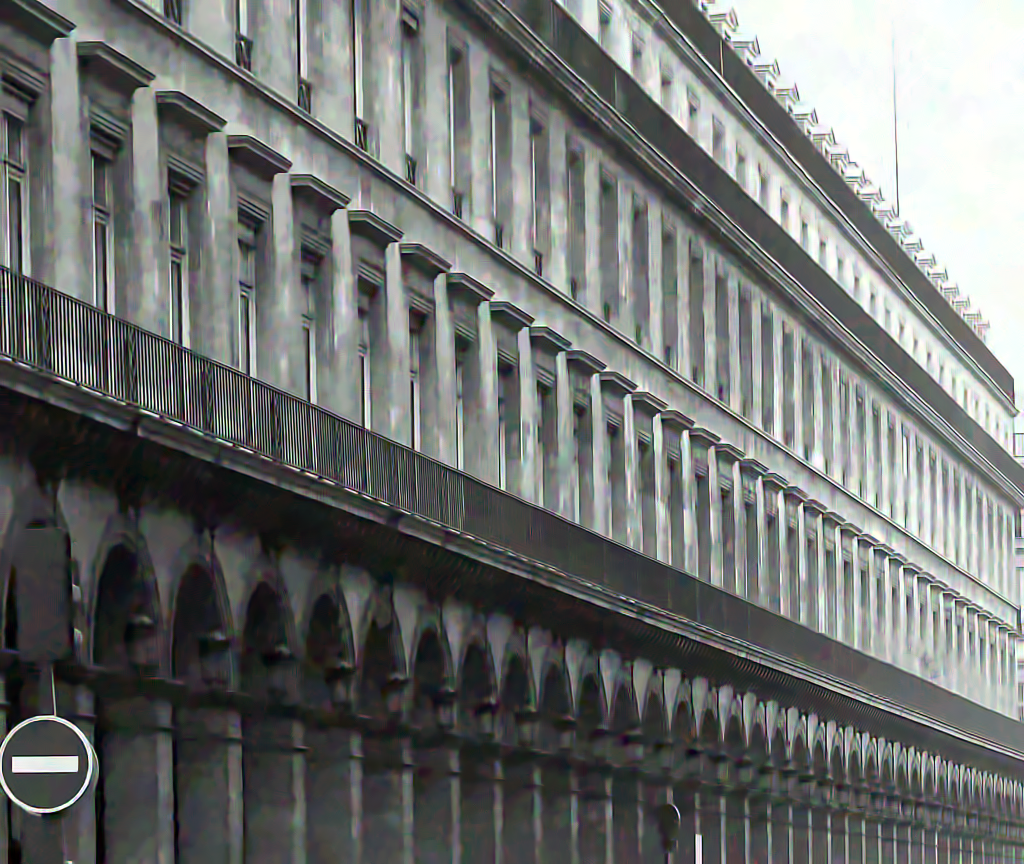} \\
\hline
PSNR &  -  &    21.22  & 22.05 & 22.34 \\ \hline \hline 

\end{tabular}
\caption{\label{Tab:image_Table} Image denoising for $\sigma = 38.25$.}
\end{center}   

\end{table}

We conclude from Table~\ref{Tab:Comparedenoise} that for a lower noise level of $\sigma = 25.5$, the unsupervised analysis dictionary learning approach Analysis K-SVD (that learns a \emph{separate} dictionary to denoise each noisy image) slightly outperforms BLORC. 
However, this is not the case for a higher noise level such as $\sigma = 38.25$. The supervised approach BLORC performs slightly better in terms of the PSNR metric for a higher level of noise.

Unsupervised dictionary learning methods like 
Analysis K-SVD rely on the structure of (unpaired) corrupted signals/images 
to learn the dictionary atoms. These structures of the signals can get buried when learning from signals with significant levels of noise. On the other hand, BLORC utilizes both the corrupted signal and the ground truth signals to learn the transform (only once on training set). 
Moreover, the transforms are learned to minimize image quality metrics of interest. 
Hence, BLORC can learn relevant features even in the presence of significant amounts of noise.

\begin{table}
\centering
\begin{tabular}{ |p{2cm}|p{2cm}|p{2cm}|p{2cm}|p{2cm}| }
\hline
  & Noisy case & BM3D & Analysis K-SVD & BLORC (ours) \\
 \hline
 \multicolumn{5}{|c|}{$\sigma$= 38.25} \\
\hline 
 \hline
 PSNR & 8.24 & 23.72 & 23.65 & $\mathbf{24.16}$ \\ 
 \hline
 \multicolumn{5}{|c|}{$\sigma$= 25.50} \\
\hline 
 \hline
 PSNR & 10.4 & 26.83 & 26.60 & 26.17\\
 \hline

\end{tabular}
\caption{Performance of different methods for image denoising (averaged over 4 images) using the PSNR (in dB) metric for two different noise levels. } \label{Tab:Comparedenoise}
\end{table}


\subsection{Comparison with other differentiable solvers}
\label{sec:compcvx}

While the BLORC algorithm uses an explicit form of the gradient,
 in automatic differentiation approaches, the task is divided into a sequence of differentiable operations as computational graphs on which backpropagation is performed through the chain rule.  This division into a sequence of operations and calculating gradients for each node of the graph can take significant time, which can be bypassed if an explicit form of the gradient already exists that connects both the upper level and the lower level problems. This advantage in time can be crucial for larger datasets and batch sizes. As a demonstration of this, we calculate $\nabla_\Wno Q$ for a  training pair $(\x_{t}, \y_{t})$ using three methods with the $\Wno = \mathbf{I}$ initialization. We note the time and accuracy for calculating a single instance of $\nabla_\Wno Q$ and average them out over 100 such different training pairs in Table~\ref{fig:imagestrips}. In the first method,  we use the direct expressions in~\eqref{eq:grad1} and~\eqref{eq:except} to get $\nabla_\Wno Q$, which we denote as "BLORC" in Table~\ref{table:compare}. For the second method, we used the CVXPY package to run an iteration of optimization over the bilevel problem for the same training pair and obtained the gradient. Finally, as the third method, we used PyTorch to calculate the gradient of \emph{our closed-form expression} in \eqref{closed-form}.

\begin{table}[hbt!]
  \centering
  \renewcommand{\arraystretch}{1.2}
  \begin{tabular}{|p{1.9cm}|c|c|c|c|}
    \hline
    \multirow{2}{1.9cm}{\textbf{Gradient Method}} & \multicolumn{2}{c|}{\textbf{$n=36$}} & \multicolumn{2}{c|}{\textbf{$n=64$}} \\
    \cline{2-5}
    & \textbf{Time (ms)} & \textbf{Error} & \textbf{Time (ms) } & \textbf{Error}\\
    \hline
    BLORC (ours) & 7.54 & 3.2e-09 &  12.65 & 5.13e-09 \\ \hline
    PyTorch  & 8.75 & 3.7e-09 & 14.03 & 5.35e-09 \ \\ \hline
    CVXPY & 17.85 & 4.8e-05  & 41.86 &  3.2e-05   \\ \hline
  \end{tabular}
   \caption{Time and error comparisons for gradient calculation averaged over 100 different single training pairs $(\x_{t}, \y_{t})$. }\label{table:compare}
\end{table}

As the baseline for our comparisons, we calculated the numerical gradient of the upper level problem by noting the incremental change in the cost for incremental changes in each element of the matrix $\Wno$. The errors for the three methods in Table~\ref{table:compare} have been calculated with the 
ground truth value being set to the one from
the numerical gradient method. The analytical form of the gradient in
BLORC makes it
faster and more accurate compared to automatic differentiation approaches as is evident from Table~\ref{table:compare}.  It is noteworthy that even when the PyTorch method uses the closed-form expression, BLORC marginally outperforms PyTorch AD method as it utilizes the analytical gradient expressions.  


\subsection{Comparison with smooth differentiable penalty} \label{sec:comp-penalty}

\textcolor{red}{
We perform an experiment to compare the transform when learned with a) an $\ell_{1}$ norm functional and with b) an element-wise Huber-loss function which is given as :}
\textcolor{red}{
\begin{align}
     L_{\delta}(x)= 
\begin{cases}
    \frac{1}{2} x^2 & \text{for } |x| < \delta \\
    \delta(|x| -\frac{1}{2}\delta),              & \text{otherwise}
\end{cases}
\end{align}
}
\textcolor{red}{
The quadratic approximation of the Huber loss function near zero makes it less effective in promoting sparsity than the classical $\ell_{1}$ norm. We perform an experiment to learn the transform using a Huber-loss penalty using a bilevel framework as follows:}

\begin{figure}[hbt!]
    \centering
     \includegraphics[width=0.7\textwidth]{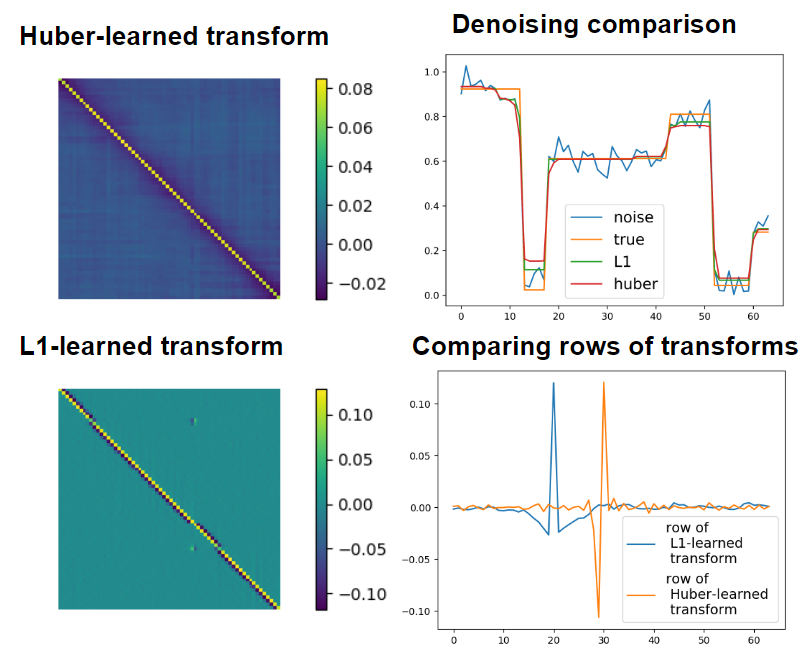}
    \caption{Comparing transforms learned with Huber penalty vs the transform learned from the $\ell_{1}$ penalty. Denoising performance was tested over a set of 10 test signals.\textcolor{blue}{The average PSNR for $\ell_{1}$ learned transform was 26.2 dB while that of Huber learned transform was 22.5 dB.}  }
    \label{fig:hubercomps}
\end{figure}

\textcolor{red}{
\begin{subequations}
\label{eq:bibeta}
\begin{align}
    & {\arg \min } \; Q(\Wno) =  \;  \frac{1}{2} \|\x_{t}^*(\Wno, \y_t) - \x_t \|_2^2  \\
   &  \textrm{s.t.} \;\, \x_{t}^*(\Wno, \y_t) = \underset{\x}{\arg \min} \frac{1}{2}\| \x - \y_t \|_{2}^2 + \sum_{i=1}^{m} L_{\delta}((\Wno \x)_{i})  \label{eq:lower-level-huber}
\end{align} 
\end{subequations}
}
\textcolor{red}{We use the autograd in CVXPY to obtain the gradient through the lower-level problem to obtain $\nabla_{\Wno} Q(\Wno)$. We perform gradient descent until convergence. The comparison between the learned transforms (note that for BLORC, we still use our direct closed-form expressions) and their denoising abilities is shown in Figure \ref{fig:hubercomps}.}

\textcolor{red}{Let the transform learned using the $\ell_{1}$ loss be denoted as $W_{\ell_{1}}$ and the transform learned using the Huber-loss functional be denoted as $W_{Hub}$. The Huber-parameter $\delta$ was optimized on a test dataset 
over a range of values in $[\frac{\|\y_{t} \|_{\infty}}{4}, \frac{3\| \y_{t} \|_{\infty}}{4}]$ and the setting achieving the best test denoising metric (PSNR) was chosen.
A careful optimization of $\delta$ is required because a higher value of $\delta$ would cause the regularizer to act more like a smooth Tikhonov regularizer and a low value of $\delta$ would make the penalty weak.} 

\textcolor{red}{
We observe from our experiments that $W_{Hub}$ (learned with careful choice of $\delta$) does not sharply promote sparsity for piecewise constant signals. This is quite evident when each row of  $W_{Hub}$ and $W_{\ell_{1}}$ are plotted together. As in Figure \ref{fig:hubercomps}, when $W_{Hub}$ is used to denoise a noisy piecewise-constant signal, it fails to reproduce the sharp edges unlike $W_{\ell_{1}}$.}

\section{Identifiability of transform under various noise levels}
\label{section:noise-levels}
 

\begin{figure}[htb]
    \centering
     \includegraphics[width=1\textwidth]{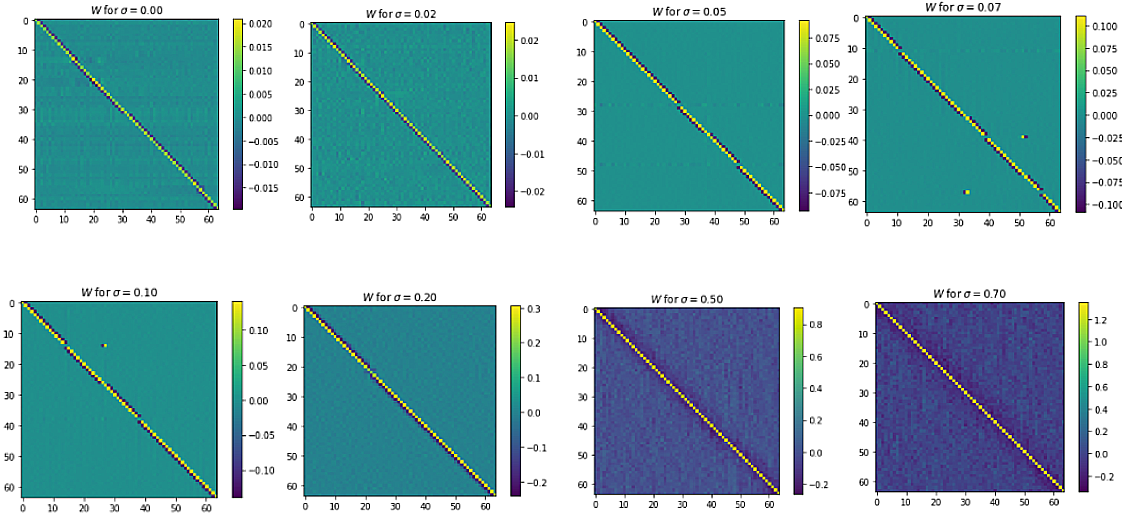}
    \caption{BLORC learned $\Wno$'s for various noise levels. In all experiments, BLORC was initialized with $\Wno = \mathbf{I}$. Supervised $(\x_{t},\y_{t})$ pairs used in the training set were piecewise-constant signals.}
    \label{fig:noiselevels}
\end{figure}

 \begin{figure}[htb]
    \centering
    \includegraphics[width=1\textwidth]{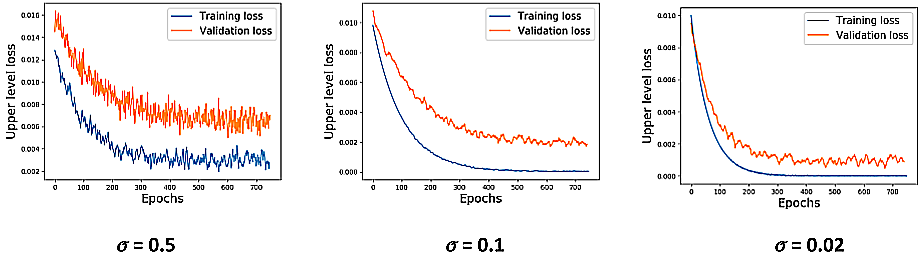}
    \caption{Training loss and validation loss vs. number of epochs for training with various noise levels. Let $\Wno_{i}$ denote the learned transform at epoch $i$. Training loss is the upper-level loss $ Q(\Wno_{i}) = \frac{1}{T} \sum_{t=1}^T \| \x_{t}^{*}(\Wno_{i},\y_{t}) -\x_{t} \|_{2}^2 $. The validation set $(\x_{l},\y_{l})_{l=1}^L$ have been generated independently from the training set but with the same noise-level. The validation loss is given as $\frac{1}{L} \sum_{l=1}^L \| \x_{l}^{*}(\Wno_{i},\y_{l}) -\x_{l} \|_{2}^2 $. Here $\x_{l}^{*}(\Wno_{i},\y_{l}) $ is the result of denoising $\y_{l} $ with learned transform at $i$th epoch $ \Wno_{i}$.  }
    \label{fig:loss}
\end{figure}


It is of interest to inspect the performance of the BLORC algorithm under various noise-levels. Also, therein lies the question: \textit{will the algorithm converge to the same the minimizer/transform for increasing noise-levels?} To answer this question, we first perform several experiments to learn the transform matrix $\Wno$ with piecewise constant training signal pairs $(\x_{t},\y_{t})_{t=1}^T$ for different noise-levels. 
Each $\x_{t}$ is a piece-wise constant signal and the noisy measurement is generated as $\y_{t} = \x_{t} +\boldsymbol{\epsilon} $, where  $\boldsymbol{\epsilon} \in \mathcal{N}(\mathbf{0},\sigma^2 \mathbf{I})$. We chose 8 noise-levels as $\sigma = 0, \, 0.02, \, 0.05, \, 0.07, \, 0.1, \, 0.2, \, 0.5, \, \text{and} \, 0.7$ ($\x_{t}$'s values are from $0$ to $1$) and performed the BLORC experiment to obtain the learned transform $\Wno$ for each of the noise levels. 
The learned transforms for the different noise levels are shown in Figure~\ref{fig:noiselevels}. For all the 8 experiments, the transform was initialized to be $\Wno =\mathbf{I}$, and we see in Figure~~\ref{fig:noiselevels} that except for $\sigma = 0.5  \ \text{and} \ 0.7$, $\Wno$ converged to something very close to the finite-differences matrix (lets call it $\Wno^*$).

Now consider the noiseless case ($\sigma = 0$), for which we have $\x_{t} =\y_{t}$. From Figure~\ref{fig:noiselevels}, we observe that the finite-difference matrix ($\Wno^{*}$) is a minimizer for the upper-level cost $Q(\Wno)$.
It is also easy to observe that $\Wno =\mathbf{0}$, the zero-matrix is a minimizer. 
In both these realizations of $\Wno$, the upper level loss ($Q(\Wno)$) reaches its minimum value, i.e., zero. 

The problem gets interesting when the noise level (standard deviation) is increased beyond $\sigma = 0$. Note that for $\Wno = \mathbf{0}$, the upper level loss is roughly $\sigma^2$, i.e., linear in the noise variance. 
This is obvious since the lower level reconstruction is just $\y_{t}$, which makes the upper level loss  $ Q(\mathbf{0}) = \frac{1}{T} \sum_{t=1}^T\|\x_{t} - \y_{t}\|_{2}^2 \approx \sigma^2$.
Thus, with higher noise levels, the zero matrix would not be expected to remain a local minimizer anymore as the upper level loss increases drastically with increase in $\sigma$.
However, we could expect $\Wno^*$ to be a minimizer even with noise.
 
It is noteworthy that beyond noise level $\sigma = 0.5$, the pattern of the learned transform $\Wno$ breaks. This is because beyond noise-level $\sigma =0.5 $, piecewise-constant structure in the corrupted signals are not preserved. 

Another interesting observation in Figure~\ref{fig:noiselevels} is that with increasing noise level, the scale of the learned transform increases (the colorbar range denotes the scale). This confirms the rather common occurrence in denoising problems. With increasing noise-level, reconstructing from the noisy signal becomes increasingly difficult, hence the regularization term is learned to take more weight than the data-fidelity term to achieve a more stable reconstruction in the lower level problem. Even at significant noise levels like $\sigma = 0.2$ (which is $20 \%$ of the peak intensity level), BLORC manages to learn a transform $\Wno$ close to the finite-differences matrix $\Wno^{*}$.

Finally,
in Figure~\ref{fig:loss}, the upper-level loss $Q(\Wno) $ is plotted for training and validation signals across training epochs. 
The validation pairs consisting of ground truth signals and noisy measurements were generated using the same noise levels as that of the training set that was used to learn the transform.
For noise levels $\sigma = 0.02$ and $0.1$, the learned transform $\Wno$ generalizes well and has a good denoising performance on unseen validation data. This is observed as the validation loss for $\sigma = 0.02 $ and $\sigma =0.1$ decreases to a significantly low value. However, for $\sigma = 0.5$, the validation loss does not decrease significantly and hence denoising using the learned transform for $\sigma = 0.5$ yields poor performance on the unseen validation set. This is also justified from the fact that in Figure~\ref{fig:loss}, the learned transform for $\sigma = 0.5$ does not resemble the structures in a finite-difference matrix.

\section{Proofs and analysis of closed-form and gradient expressions} 
\label{section:derivations}
In this section, we present the derivations of the closed-form expressions and gradients and an analysis of the closed-form expression on Stiefel manifolds. First, we present a proof for Theorem~\ref{closed-form}.
\subsection{Obtaining the closed-form: Proof for Theorem ~\ref{closed-form}} \label{obtain_close}
We propose to solve signal reconstruction problems using a sparsifying analysis operator that is learned from training data in a supervised way.
Our learning problem is 
\begin{subequations}
\begin{equation}
    \argmin_\Wno \Q(\Wno), \text{\quad where \quad}
    \Q(\Wno) = \frac{1}{T}\sum_{t=1}^T \frac{1}{2} \|\x_{t}^*(\W, \y_t) - \x_t \|_2^2,
    \label{eq:main_upper}
\end{equation}
where
$\x_t$ and $\y_t \in \mathbb{R}^n$ are the $t$th training signal and its corresponding noise-corrupted version, $\W \in \mathbb{R}^{n \times n}$ is a sparsifying operator we intend to learn with the help of the corresponding lower level problem,
  \begin{equation}  
   \quad \x_{t}^*(\Wno, \y_{t}) = \argmin_\x 
    \frac{1}{2}\| \x - \y_{t} \|_2^2 + \beta \| \Wno \x \|_1.
    \label{eq:main_lower}
\end{equation}
\label{eq:main}
\end{subequations}

Let $\vec{c}(\Wno)$ denote the sign pattern
associated with a given $\Wno$,
i.e.,
$\vec{c}(\Wno) = \sign(\Wno \x^*(\Wno))$,
where $[\sign(\z)]_i$ 
is defined to be -1 when $[\z]_i < 0$;
0 when $[\z]_i = 0$;
and 1 when $[\z]_i > 0$. Note that we omit the $t$-dependence from $\x^*(\Wno)$ as it is not relevant for the derivation. 
Considering a fixed sign-pattern $\vec{c}_0 =\vec{c}(\Wno)$,
we define matrices that pull out the rows
of $\Wno$ that give rise to zero, negative, and positive values in $\Wno \x^*(\Wno)$.
Let $k_{=0}$, $k_{\ne0}$, $k_{<0}$, and $k_{>0}$ denote the number of
zero, nonzero, negative, and positive elements of $\vec{c}$, respectively.
Similarly, let $[\vec{\pi}_0]_m$, $[\vec{\pi}_{<0}]_m$, and $[\vec{\pi}_{>0}]_m$
denote the indices of the $m$th zero, negative, and positive element of $\vec{c}_0$, respectively.
Let $\Szero \in \mathbb{R}^{k_{=0} \times k}$ and $\Spm \in \mathbb{R}^{k_{\ne0} \times k}$
be defined as
\begin{equation}
    [\Szero]_{m,n}=  \begin{cases}
      1 &\quad \text{if} \quad [\pi_0]_m = n;\\
      0 &\quad \text{otherwise};
\end{cases}
\text{\quad and \quad}
    [\Spm]_{m,n}=  \begin{cases}
      1 &\quad \text{if} \quad [\pi_{>0}]_m = n;\\
      -1 &\quad \text{if} \quad [\pi_{<0}]_{m-k_{>0}} = n;\\
      0 &\quad \text{otherwise.}
\end{cases}
\end{equation}

The nonzero sign pattern is $\s = \Spm \vec{c}_0$. Let $\Wzero = \Szero \Wno$ and $\Wpm =\Spm \Wno$ contain the rows of $\Wno$, whose indices are given by the sets $k_{=0}$ and $k_{\ne0}$, respectively.
With these notations in place, we can write 
\begin{equation}
  \label{eq:linear-transform-constrained}
  \x^*(\Wno) = 
  \argmin_{\vec{x}} \frac{1}{2}\| \vec{x} - \vec{y} \|_2^2 + 
  \beta \|\Wno \vec{x} \|_1, \quad
   \text{s.t.} \quad \Wzero\x = \vec{0}
\end{equation}
for all $\Wno$ such that $\vec{c}(\Wno) = \vec{c}_0$.
This is true because 
whenever $\vec{c}(\Wno) = \vec{c}_0$,
the minimizer of \eqref{eq:analysis}
is feasible for \eqref{eq:linear-transform-constrained}.
Similarly,
we use $\Wpm$ to simplify the $\ell_1$ norm term,
\begin{equation}
  \label{eq:linear-transform-local}
  \x^*(\Wno) = 
  \argmin_{\vec{x}} \frac{1}{2}\| \vec{x} - \vec{y} \|_2^2 
  + \beta \s\tran \Wpm \vec{x}, \quad
   \text{s.t.} \quad \Wzero\x = \vec{0},
\end{equation}
which holds again for all $\Wno$ such that $\vec{c}(\Wno) = \vec{c}_0$.

Now that we have transformed the problem into an equality-constrained quadratic minimization,
we can use standard results  
(e.g., see \cite{boyd_convex_2004}, Section 10.1.1)
to state the KKT conditions for \eqref{eq:linear-transform-local} as
 \begin{equation}
  \label{eq:linear-transform-KKT}
  \underbrace{
  \begin{bmatrix}
    \vec{I} & \Wzero\tran \\
    \Wzero & \vec{0}
  \end{bmatrix}}_{\vec{A} \in \mathbb{R}^{(n + k_{=0}) \times (n + k_{=0})}}
\underbrace{
  \begin{bmatrix}
    \x^*(\Wno) \\
    \vec{\nu}
  \end{bmatrix}}_{\vec{z} \in \mathbb{R}^{n + k_{=0}}}
=
\underbrace{
  \begin{bmatrix}
    \y - (\beta \s\tran\Wpm)\tran \\
    \vec{0}
  \end{bmatrix}}_{\vec{b} \in \mathbb{R}^{n+k_{=0}}},
  \end{equation}
where the underbraces give names ($\vec{A}$, $\vec{z}$, and $\vec{b}$) to each quantity
to simplify the subsequent notation.
Because $\vec{I}$ is nonsingular,
$\vec{A}$ is invertible whenever $ \Wzero$ has full row rank~\cite{boyd_convex_2004}.
Lastly, in order to extract the part corresponding to $x^*(\Wno)$, we define a selection matrix  $\vec{P}_\x \in \mathbb{R}^{n \times (n + k_{=0})}$ defined to be $\vec{P}_\x =
  \begin{bmatrix}
  \vec{I} & \vec{0} \end{bmatrix}$. The part corresponding to $x^*(\Wno)$ can be extracted as $\x^*(\Wno) = \vec{P}_\x \vec{A}^{-1} \vec{b}$. If $ \Wzero$ has full row rank, then by block matrix inversion formula corresponding to the first term we have the final closed-form expression: 
  
  \begin{lemma}
  \label{blockmat}
  Given a block matrix partitioned into four blocks, it can be inverted blockwise in the following manner: If $\mathbf{P} = \begin{bmatrix}
     \mathbf{L} &  \mathbf{M}\\
     \mathbf{N} &  \mathbf{O}
  \end{bmatrix}$ \\ where $\mathbf{L}$ and $\mathbf{O}$ are arbitrary sized, and $\mathbf{M}$ and $\mathbf{N}$ are conformable for partitioning.
  then $\mathbf{P}^{-1} =  \begin{bmatrix}
     \mathbf{L} + \mathbf{L}^{-1}\mathbf{M}(\mathbf{O} - \mathbf{N}\mathbf{L}^{-1}\mathbf{M})^{-1}\mathbf{N}\mathbf{L}^{-1} &   -\mathbf{L}^{-1}\mathbf{M}(\mathbf{O} - \mathbf{N}\mathbf{L}^{-1}\mathbf{M})^{-1}\\
     -(\mathbf{O} - \mathbf{N}\mathbf{L}^{-1}\mathbf{M})^{-1}\mathbf{N}\mathbf{L}^{-1} &  (\mathbf{O} - \mathbf{N}\mathbf{L}^{-1}\mathbf{M})^{-1}
  \end{bmatrix}$ \cite{bernstein2009matrix}. Furthermore, $\mathbf{L}$ and its schur complement $(\mathbf{O} - \mathbf{N}\mathbf{L}^{-1}\mathbf{M})$ must be invertible. 
  \end{lemma}

Now, according to Lemma~\ref{blockmat} and considering the block matrix $\mathbf{A} =\begin{bmatrix}
    \vec{I} & \Wzero\tran \\
    \Wzero & \vec{0}
  \end{bmatrix}$, the condition for $\mathbf{A}$ to be invertible here is for $ -\Wzero \Wzero\trans$ (Schur complement of $\mathbf{A}$) to be invertible. 
  The necessary and sufficient condition for $\Wzero \Wzero\trans$ to be invertible is for $\Wzero$ to have a full row-rank. 
  Now, extracting the term corresponding to $\x^*(\Wno) $ from $\vec{z}$ using the block matrix inversion formula in Lemma~ \ref{blockmat}, we would have:
  
 \begin{align*}
    &  \x^*(\Wno) = (\mathbf{I} +  \mathbf{I}^{-1}\mathbf{\Wzero}\tran(\mathsf{0}-\mathbf{\Wzero} \mathbf{I}^{-1}\mathbf{\Wzero}\tran)^{-1}\mathbf{\Wzero}\mathbf{I}^{-1}) (\y - \beta \Wpm\tran\s)  \\
        &\;\;\;\;\; \;\;\;\;\;\; =  (\mathbf{I} - \mathbf{\Wzero}\tran(\mathbf{\Wzero}\mathbf{\Wzero}\tran)^{-1}\mathbf{\Wzero})(\y - \beta \Wpm\tran\s)  \\ 
        & \implies \boxed{ \x^*(\Wno)= \proj_\nullspace{\Wzero} (\y - \beta\Wpm \tran \s ).} 
 \end{align*}
Here, $\proj_\nullspace{\Wzero}$ denotes the projection operator onto the null-space of $\Wzero$. 
  
We also observe that this similar closed-form expression has also been obtained in Equation (33) of \cite{tibshirani2011solution} by formulating the dual problem.

Also, it is noteworthy that the closed-form expression derived above holds even when $\Wzero$ is not full row-rank but with a slight modification as follows. 
Note that when  $\Wzero$ is not full row-rank, $\vec{A}$ defined in \eqref{eq:linear-transform-KKT} is also rank-deficient. Hence to calculate the closed-form $\x^{*}(\Wno)$, we use the expression $ \x^*(\Wno) = \vec{P}_\x \vec{A}^{\dagger} \vec{b} $, where $\vec{A}^{\dagger}$ denotes the pseudo-inverse of $\vec{A}$. It is then, straightforward to prove that the updated closed-form expression is given as $ \x^{*}(\Wno) =  (\mathbf{I} - \mathbf{\Wzero}\tran(\mathbf{\Wzero}\mathbf{\Wzero}\tran)^{\dagger}\mathbf{\Wzero})(\y - \beta \Wpm\tran\s) $, where $\dagger$ denotes the pseudo-inverse operator. This can be derived from the formula of psuedo-inverse of block-matrices as given in \cite{hung1975moore}. The term $(\mathbf{\Wzero}\mathbf{\Wzero}\tran)^{\dagger} $ becomes $(\mathbf{\Wzero}\mathbf{\Wzero}\tran)^{-1} $
 only when $\Wzero$ is full row-rank which is exactly the boxed expression above.

Next, we present the derivations of the gradient of the closed-form expression that we stated in Corollary \ref{corollary}.

\subsection{Gradients of the closed-form: Proof for Corollary \ref{corollary}}
\label{grad_close}
Now that we have got a closed-form expression for the lower level cost, we can rewrite the original bilevel optimization problem~\eqref{eq:bilevel} as
\begin{equation}
    \argmin_{\Wno} Q(\Wno), \text{\quad where \quad}
    Q(\Wno ) = \frac{1}{T}\sum_{t=1}^T \frac{1}{2} \|\x^*(\Wno, \y_t) - \x_t \|_2^2,
    \end{equation}
  \begin{equation}  
  \text{s.t. \quad} \x^*(\Wno,\y_t)= (I- \proj_{row(\Wzero)})(\y_t - \beta \Wpm \tran s )
  \label{eq:denoising_closed}
\end{equation}%

Since we perform SGD based updates, we calculate the gradient of the upper level cost function for each training pair and take a gradient step. Let $Q(\Wno) =  \frac{1}{2T} \|\x^*(\Wno, \y_t) - \x_t \|_2^2$ be the upper level cost function for each training pair, then $\partial Q = \nabla_{\x} Q \tran \partial \x^* $.

Next, we aim to differentiate \eqref{eq:denoising_closed} with respect to the elements of $\Wno$.
 We proceed by differentiating separately with respect to $\Wpm$ and $\Wzero$ because they form a partition of the rows of $\W$.
 The $\Wpm$ part follows from simple matrix calculus rules~\cite{minka_old_2000} as
\begin{equation}
    \partial \xopt =  -\beta \proj_\nullspace{\Wzero}  \partial \Wpm\trans \s.
    \label{eq:expr1}
\end{equation}
The $\Wzero$ part requires differentiation through a null space projection,
for which we rely on Theorem~4.3 in~\cite{golub_differentiation_1973}.
The result is 

\begin{equation} 
\label{eq:dxdWzero}
\partial \xopt  = -(\Wzero^\dagger \partial\Wzero \proj_\nullspace{\Wzero}
+  (\Wzero^\dagger \partial \Wzero \proj_\nullspace{\Wzero})\trans ) (\y - \beta \Wpm\trans \s), 
\end{equation} 

Finally,
we are interested in computing the gradient with respect to $\W$ of an upper-level cost functional,
$Q : \mathbb{R}^{n \times n} \to \mathbb{R}$;
this can be achieved via the chain rule
(writing $\partial Q = \nabla_\xopt Q\trans \partial \xopt$, substituting $\partial\xopt$,
and rearranging into canonical form~\cite{minka_old_2000}),
which results in
\begin{align}
    & \nabla_{\Wpm} Q = -\beta \s \nabla_{\xopt} Q\trans    \proj_\nullspace{\Wzero}\\
    & \nabla_{\Wzero} Q= -  (\proj_\nullspace{\Wzero} 
    (\vec{q} \nabla_{\xopt} Q \trans  + \nabla_{\xopt} Q \vec{q}\trans)
    \Wzero^+)\trans
\end{align}
with $\vec{q} = \y - \beta \Wpm\trans \s$.
Here, we choose $Q(\Wno) = \frac{1}{2T}\|\xopt(\Wno) - \x_{t}\|_{2}^2$,
so $\nabla_\xopt Q = \frac{1}{T}(\xopt - \x_{t})$.

\begin{proof}
Given the expression $\partial Q = \nabla_{\x^*} Q \tran \partial \x^* $, we replace $\partial \x^*$ with \eqref{eq:expr1} and \eqref{eq:dxdWzero}, to obtain the gradient of the upper level cost function w.r.t. elements of the $\Wno$ matrix (more specifically, w.r.t. $\Wzero$ and $\Wpm$). 

First, we replace $\partial \x^{*}$ with Equation~\eqref{eq:expr1} in order to have a relation between $\partial Q$ and $\partial \Wpm$. 

\begin{align*}
   &  \partial Q = \nabla_{\x^{*}} Q \tran \partial \x^*= -\beta \Tr (\nabla_{\x^{*}} Q   \tran  \proj_\nullspace{\Wzero} \partial \Wpm\trans \s)  \\
   &  = -\beta \Tr ( \s \nabla_{\x^{*}} Q   \tran \proj_\nullspace{\Wzero} \partial \Wpm\trans ) &  \text{\quad [$\Tr(\mathbf{ABC}) =\Tr(\mathbf{CAB})$] \quad}  \\
   & = -\beta \Tr ( (\s \nabla_{\x^{*}} Q   \tran \proj_\nullspace{\Wzero}) \tran \partial \Wpm ). &  \text{\quad [$\Tr(\mathbf{A B \trans}) =\Tr(\mathbf{B A \trans})$] \quad}
\end{align*}

Recalling that in Section~\ref{Grad_cal}, if the differential of a scalar and matrix is related as $\partial Q = \Tr(\E^T\partial \Wpm)$, then $\nabla_{\Wpm}Q = \E$. Hence, we have from the last line, $\nabla_{\Wpm}Q = -\beta (\s \nabla_{\x^{*}} Q   \tran \proj_\nullspace{\Wzero}) $.

Next, replacing $\partial \x^{*}$ with \eqref{eq:dxdWzero}, we can similarly have a relation between $\partial Q$ and $\partial \Wzero$.
\begin{align*} 
    & \partial Q = \nabla_{\x^{*}} Q \tran \partial \x^* = \Tr( \nabla_{\x^{*}} Q \tran \partial \x^*) \\
    & =  - \Tr(\nabla_{\x^{*}} Q \tran (\Wzero^+ \partial\Wzero \proj_\nullspace{\Wzero}
+  (\Wzero^+ \partial \Wzero \proj_\nullspace{\Wzero})\trans ) \q) \\
    & = - \Tr(\nabla_{\x^{*}} Q \tran \Wzero^+ \partial\Wzero \proj_\nullspace{\Wzero} \q ) - \underbrace{ \Tr(\nabla_{\x^{*}} Q \tran (\Wzero^+ \partial\Wzero \proj_\nullspace{\Wzero}) \trans \q)}_{\text{Apply} \quad \Tr( \mathbf{A \tran B\tran C}) = \Tr(\mathbf{ B A C \tran)} } \\
    & =  - \underbrace{\Tr(\nabla_{\x^{*}} Q \tran \Wzero^+ \partial\Wzero \proj_\nullspace{\Wzero} \q )}_{\text{Apply} \quad \Tr(\mathbf{ABC}) =\Tr(\mathbf{CAB})  } - \underbrace{\Tr((\Wzero^+ \partial\Wzero \proj_\nullspace{\Wzero}) \nabla_{\x^{*}} Q  \q \tran)}_{\text{Apply} \quad \Tr(\mathbf{ABC}) =\Tr(\mathbf{CAB})  }  \\
    & =  - \Tr(  \proj_\nullspace{\Wzero} \q \nabla_{\x^{*}} Q \tran \Wzero^+ \partial\Wzero) -
    \Tr( \proj_\nullspace{\Wzero} \nabla_{\x^{*}} Q   \q \tran \Wzero^+ \partial\Wzero)\\
    & = - \Tr(\proj_\nullspace{\Wzero} (\q \nabla_{\x^{*}} Q \tran  +  \nabla_{\x^{*}} Q \q\tran  ) \Wzero^+ \partial\Wzero )
\end{align*}

We see now that the differentials are related by the form $\partial Q = \Tr(\E^T \partial \Wzero) $, hence we directly have $\nabla_{\Wzero} Q =  -  (\proj_\nullspace{\Wzero} 
    (\vec{q} \nabla_{\xopt} Q \trans  + \nabla_{\xopt} Q \vec{q}\trans)
    \Wzero^+)\trans  $.

\end{proof}

\subsection{Analysis of closed-form expression on the Stiefel manifold}
\label{section:Stiefel}
This section analyses the closed-form expression for the denoising problem obtained on the Stiefel manifold (under the orthogonal constraint) $\Wno \Wno^T =\mathbf{I}$. \textcolor{red}{In this case, gradients from the upper-level can be projected to the Stiefel manifold using smooth retraction \cite{birtea2019first}.} The non-differentiable points of the closed-form expression 
$\x_t^*(\Wno) = \Wno^T \mathcal{S}_{\lambda}(\Wno \y_t)$ are given by $[\Wno \y_t]_{i} = \pm \lambda$, i.e., an inner-product of a transform row with the measurement vector is either $\lambda$ or $-\lambda$. We show that these non-differentiable points form a set of measure zero on the Stiefel manifold.

\begin{wrapfigure}{l}{0.5\textwidth}
  \begin{center}
     \includegraphics[width=0.48\textwidth]{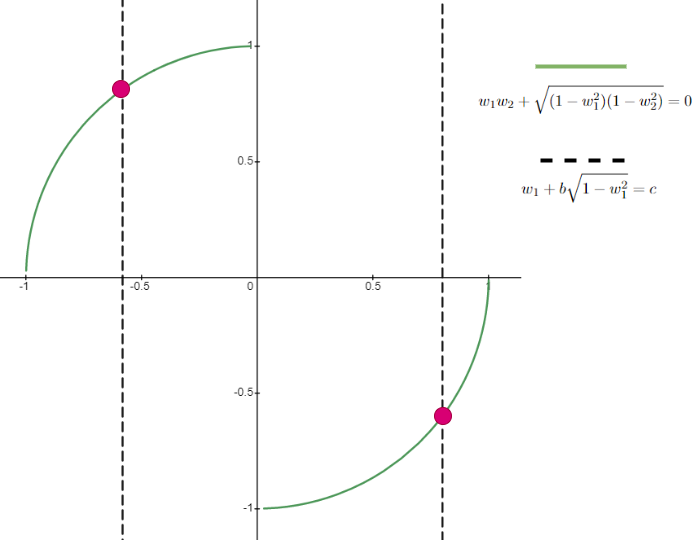}
  \end{center}
  \caption{Non-differentiable points (red) of the closed-form expression $\x_t^*(\Wno) = \Wno^T \mathcal{S}_{\lambda}(\Wno \y_t)$ on the Stiefel manifold (green) for matrices in $\mathbb{R}^{2\times 2}$, shown in the parameter space $ [w_{1},w_{2}]$.}
  \label{fig:1dcase}
\end{wrapfigure}

For simplicity, we start by considering a matrix in $\mathbb{R}^{2 \times 2}$, which also lies on the Stiefel manifold. Such a matrix would take the form $\Wno = \begin{bmatrix}
w_{1} & \sqrt{1-w_{1}^2}\\
w_{2} & \sqrt{1-w_{2}^2}
\end{bmatrix}$, as the rows are normalized. 
Due to the orthogonality condition on the rows, we have $w_{1}w_{2} + \sqrt{(1-w_{1}^2)(1-w_{2}^2)} = 0$. The locus of the curve forms the Stiefel manifold in the domain of $ [w_{1},w_{2}] \in \mathbb{R}^{2}$, and is shown by the green circular curve in Figure~\ref{fig:1dcase}. 
The non-differentiable points lie on $[\Wno \y]_{i} = \pm \lambda$ (for a generic $\y$). Without loss of generality, we assume the point of non-differentiability holds only on the first index of the vector, i.e., $[\Wno \y]_{1} = \pm \lambda$.
It is easy to see that in $\mathbb{R}^2$, the non-differentiablity points of the closed-form expression would lie on the curve $ w_{1} + b \sqrt{1 - w_{1}^2} = c$, as denoted by black dashed lines in Figure~\ref{fig:1dcase}.
These lines intersect the manifold curve (shown in green) only at finite points as shown by the red blobs.  The dimension of the Stiefel manifold for matrices in $ \mathbb{R}^{2\times 2} $ is 1\footnote{The green curve can be parameterized by $[\sin(\theta),\cos(\theta)]$ and hence has only a single degree of freedom, i.e., on $\theta$, where $\theta \in {[\frac{\pi}{2},\pi] \cup [\frac{3\pi}{2},2\pi]}$. The angle restriction is to maintain row-orthogonality.}, whereas the dimension of the non-differentiable points on the manifold (here red points) is $0$. Then, according to Theorem 2.22 in~\cite{rudin1970real}, the non-differentiable points form a set of Lebesgue measure $0$ as this intersection has a dimension lower than the dimension of the manifold. 

Next, we provide a brief discussion for matrices in $\mathbb{R}^{n \times n}$ lying on the Stiefel manifold ($V_{n}(\mathbb{R}^{n})$), or the set of orthogonal $n$-frames in  $\mathbb{R}^{n} $.
A non-constrained matrix in $\mathbb{R}^{n \times n}$ has dimension of $n^2$ as there are $n^2$ free parameters. Now as each row is normalized, the parameters in each row are constrained by the equation $\sum_{j=1}^n w^2_{ij}=1$. 
Hence, for $n$ rows, the degrees of freedom decrease by $n$. Moreover, as the rows are orthogonal to each other, 
each (independent)
orthogonality constraint is captured by $\sum_{j=1}^n w_{ij}w_{kj}=0 \text{ for $ k \neq i$ }$. As there are $n \choose 2$ distinct pairs of rows, the degrees of freedom reduce by $n \choose 2 $ from the orthogonality conditions. Hence, the effective dimension of a Stiefel manifold for matrices in $\mathbb{R}^{n\times n}$ could be seen as $n^2 - n -{n \choose 2} = \frac{n(n-1)}{2}$. 

Now without loss of generality\footnote{If the non-differentiability holds in other indices as well, that does not increase the dimension of the set in which these points lie. Hence, the argument still holds.}, consider that the closed-form expression is not differentiable only in the first index, i.e., $[\Wno \y ]_{1}=\lambda $ (could also be $-\lambda$ equivalently). So the variables in the first row of $\Wno$ are constrained by an extra equation. This decreases the overall degrees of freedom by one and hence the effective dimension of the intersection of the hyperplane $[\Wno \y ]_{1}=\lambda $ ($\lambda>0$) and the Stiefel manifold is $\frac{n(n-1)}{2} -1$, which is one less than the dimension of the manifold itself. Thus, these non-differentiable points would form a set of measure zero on the Stiefel manifold. 


\section{Conclusions}
\label{sec:conclusions}
This paper presented an approach for supervised learning of sparsity-promoting nonsmooth ($\ell_1$) regularizers for denoising problems.
The underlying training problem is a challenging bilevel optimization problem, where the upper level loss is a task-based one (e.g., mean squared error for denoising) and the lower level problem is a variational problem for denoising, whose solution is used in the upper level loss.
Our approach, BLORC, allows learning the sparsifying operator in the lower level problem  by exploiting local closed-form expressions for the solution of the variational problem.
The closed-form expressions enable computing gradients directly and efficiently with respect to the operator parameters.
Experimental results show the ability of BLORC to learn underlying sparsifying operators robustly in the presence of noise for both 1D signals and 2D images.
For images, the proposed approach outperforms recent denoising schemes, including unsupervised analysis dictionary learning.
\textcolor{red}{In future work, we plan to extend the approach to handle non-unique lower-level minimizers and extend to
other inverse problems with complex-valued images such as in magnetic resonance imaging. }

\section*{Acknowledgments}
We thank Jeffrey Fessler and Caroline Crockett, University of Michigan, for helpful discussions and their comments on this work.


\bibliographystyle{siamplain}
\bibliography{SIIMS}

\end{document}